\pdfoutput=1
\documentclass[11pt]{article}

\usepackage[final]{acl}
\usepackage{xcolor}
\usepackage{times}
\usepackage{latexsym}
\usepackage[T1]{fontenc}
\usepackage[utf8]{inputenc}

\usepackage{microtype}

\usepackage{inconsolata}
\usepackage{amsmath}
\usepackage{graphicx}
\graphicspath{{img/}}
\title{GenPilot: A Multi-Agent System for Test-Time Prompt Optimization in Image Generation}

\author{
 \textbf{Wen Ye\textsuperscript{1,2}},
 \textbf{Zhaocheng Liu\textsuperscript{3}},
 \textbf{Yuwei Gui\textsuperscript{6}},
 \textbf{Tingyu Yuan\textsuperscript{4,2}},
  \textbf{Yunyue Su\textsuperscript{1}},
   \textbf{Bowen Fang\textsuperscript{1,2}}\\
    \textbf{Chaoyang Zhao\textsuperscript{4,5}},
        \textbf{Qiang Liu\textsuperscript{1}},
        \textbf{Liang Wang\textsuperscript{1}\thanks{Corresponding author.}}
\\
 \textsuperscript{1}New Laboratory of Pattern Recognition (NLPR), 
 Institute of Automation, \\ Chinese Academy of Sciences (CASIA) 
 \textsuperscript{2}School of Artificial Intelligence, \\
 University of Chinese Academy of Sciences 
 \textsuperscript{3} Baichuan Inc.\\
 \textsuperscript{4}Foundation Model Research Center, 
 Institute of Automation, \\ Chinese Academy of Sciences (CASIA) 
\textsuperscript{5}Objecteye.Inc \\
 \textsuperscript{6}Beijing University of Posts and Telecommunications
\\
 \texttt{yewen2025@ia.ac.cn},
 \texttt{lio.h.zen@gmail.com},
 \texttt{guiyuwei@bupt.edu.cn}\\
 \texttt{\{yuantingyu2024, yunyue.su,\}@ia.ac.cn},
  \texttt{bwn.fang@gmail.com}\\
 \texttt{\{chaoyang.zhao,qiang.liu,wangliang\}@nlpr.ia.ac.cn}
\\
}

\begin{document}
\maketitle
\begin{abstract}
Text-to-image synthesis has made remarkable progress, yet accurately interpreting complex and lengthy prompts remains challenging, often resulting in semantic inconsistencies and missing details.
Existing solutions, such as fine-tuning, are model-specific and require training, while prior automatic prompt optimization (APO) approaches typically lack systematic error analysis and refinement strategies, resulting in limited reliability and effectiveness.
Meanwhile, test-time scaling methods operate on fixed prompts and on noise or sample numbers, limiting their interpretability and adaptability.
To solve these, we introduce a flexible and efficient test-time prompt optimization strategy that operates directly on the input text.
% enabling better interpretability and broader model compatibility. 
%
We propose a plug-and-play multi-agent system called GenPilot, integrating error analysis, clustering-based adaptive exploration, fine-grained verification, and a memory module for iterative optimization.
Our approach is model-agnostic, interpretable, and well-suited for handling long and complex prompts.
Simultaneously, we summarize the common patterns of errors and the refinement strategy, offering more experience and encouraging further exploration.
Experiments on DPG-bench and Geneval with improvements of up to 16.9\% and 5.7\% demonstrate the strong capability of our methods in enhancing the text and image consistency and structural coherence of generated images, revealing the effectiveness of our test-time prompt optimization strategy.
The code is available at \url{https://github.com/27yw/GenPilot}.
\end{abstract}

\section{Introduction}
Recently, text-to-image generation models \cite{ho2020denoisingdiffusionprobabilisticmodels, rombach2022highresolutionimagesynthesislatent, ramesh2022hierarchical} have witnessed remarkable developments, indicating their excellent performance across a multitude of applications.
Nevertheless, translating complex and compositional prompts into semantically aligned, high-fidelity images remains a significant challenge.
As prompt complexity increases, existing models struggle to preserve semantic coherence, exposing a persistent semantic gap and resulting in compositionality catastrophe.
These limitations are further exacerbated by architectural inconsistencies across models, which hinder the development of a unified and generalizable framework adaptable to diverse T2I paradigms.

To improve multimodal alignment in T2I generation, existing efforts \cite{mañas2024improvingtexttoimageconsistencyautomatic, fu2024ap, saharia2022photorealistictexttoimagediffusionmodels} can be broadly categorized into fine-tuning and prompting.
While fine-tuning or retraining model parameters to capture detailed semantics information, it is often computationally intensive and model-specific.
In contrast, manual prompting relies heavily on human intuition, lacking scalability across prompts, tasks, and architectures.
Recent works, such as OPT2I~\cite{mañas2024improvingtexttoimageconsistencyautomatic}, DPO-Diff~\cite{wang2024discretepromptoptimizationdiffusion}, and AP-Adapter~\cite{fu2024ap}, explore automatic prompt optimization to enhance generation quality.
However, most approaches require additional training and are designed for certain models, also often lack systematic error analysis.
\begin{figure*}[ht!]
\begin{center}
\includegraphics[width=\linewidth]{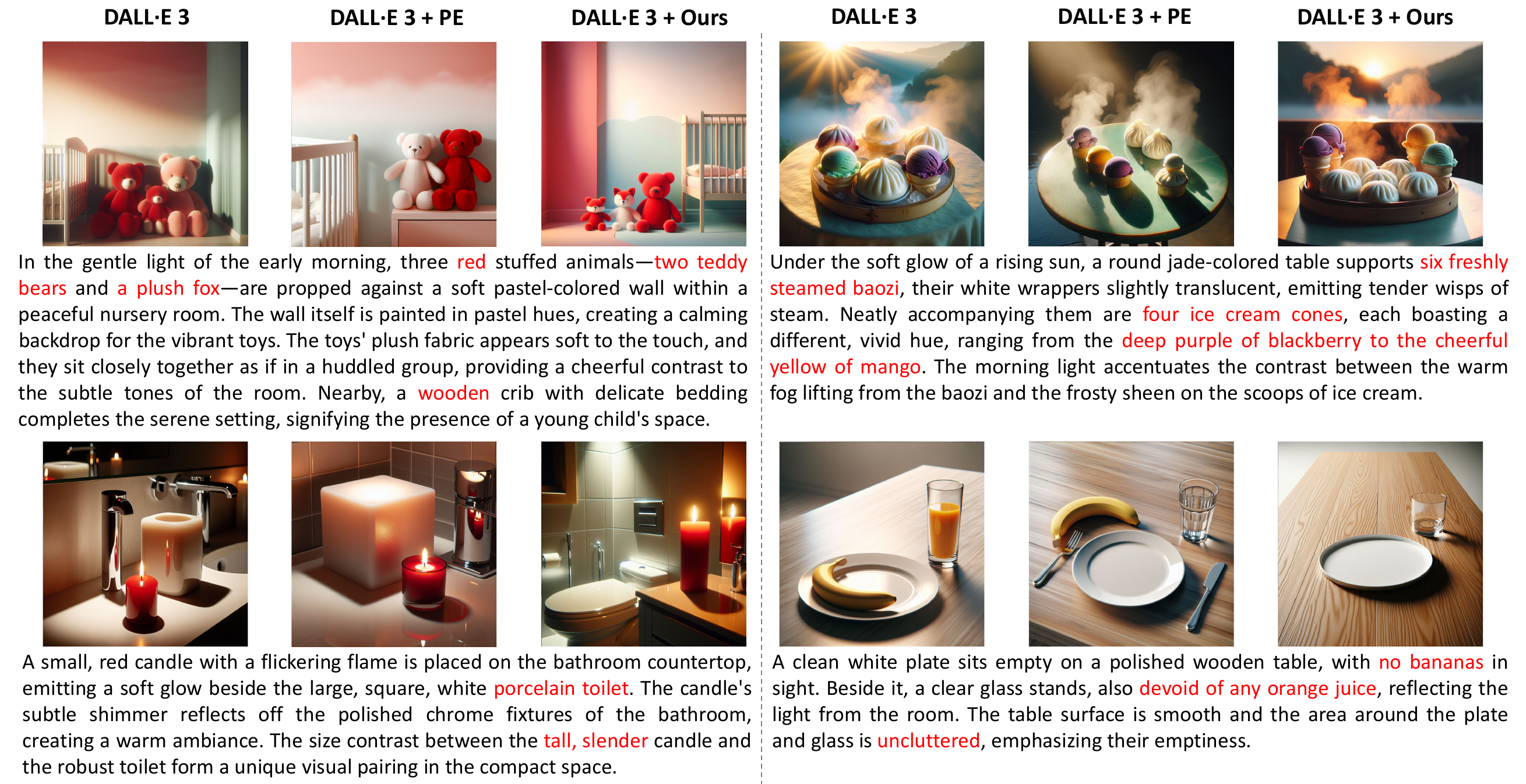}
\end{center}
\caption{
Visualized examples from DALL-E 3 \cite{betker2023improving} with GenPilot processing complicated and lengthy prompts.
Compared to the prompt engineering (PE), generative models with GenPilot successfully achieve accurate results, addressing both the semantic gap and even the challenging tasks of exclusion of certain objects.
}\label{fig:show_case_5}
\end{figure*}
With the advancement of large language models, test-time scaling has been explored in various scenarios by leveraging additional computational resources and inference-time adjustments to improve performance.
%
% Some studies extend this idea to image generation, for instance, a novel inference-time scaling framework for diffusion models \cite{ma2025inferencetimescalingdiffusionmodels} formulates the sampling process as a search problem over noise space with a verifier for evaluation.
% %
Some studies extend this idea to image generation. SANA-1.5 \cite{xie2025sana15efficientscaling} generates many samples and a verifier selects the best sample.

Although recent progress in automatic prompt optimization (APO) and test-time scaling (TTS) has improved image generation, they still suffer from limitations such as reliance on random exploration or fixed prompts, lack of systematic error identification, or coarse-grained verification, hindering flexibility and interoperability.
% APO often lacks systematic error identification and relies on random exploration or model training, while TTS typically uses fixed prompts and operates on noise space with coarse-grained verifiers, limiting flexibility and interpretability. 
%
To address these, we propose GenPilot, a plug-and-play multi-agent system that brings test-time scaling into the prompt space by formulating the prompt optimization as a search problem, enabling dynamic and interpretable prompt refinement.
GenPilot is broadly applicable across diverse models without model training to improve the prompts for image generation.
Examples are presented in Figure \ref{fig:show_case_5}.

Our system contains two main stages: the error analysis module and the test-time prompt optimization module.
In Error Analysis, GenPilot decomposes the initial prompt, leverages visual question answering (VQA) and captioning to detect and localize semantic inconsistencies.
During test-time optimization, GenPilot iteratively refines the prompt based on errors and memory feedback with a multi-modal large language model (MLLM) \cite{bai2025qwen25vltechnicalreport} scorer, cluster, and memory.

The main contributions are three-fold:

\begin{itemize}
    \item 
We propose \textbf{GenPilot}, a plug-and-play multi-agent system that performs test-time prompt optimization as a search problem for interpretable results, improving image consistency without training across diverse T2I models.

    \item 
GenPilot introduces systematic error analysis and fine-grained verification, enabling dynamic prompt exploration via clustering and iterative feedback, and memory updates.
    \item 
    Experiments on both long prompts from DPG-bench \cite{hu2024ellaequipdiffusionmodels} and short prompts from Geneval \cite{ghosh2023genevalobjectfocusedframeworkevaluating} show that GenPilot consistently improves performance across models, demonstrating robustness and generalizability for T2I tasks.
\end{itemize}

\section{Related Work}
\subsection{Text to Image Generation}
Recently, text-to-image models (T2I models) have developed rapidly.
Nonetheless, their performance is restricted not only by architectural design but also by the quality of the input prompts.
Early methods such as Stable Diffusion models (SD) \cite{rombach2022highresolutionimagesynthesislatent} rely on CLIP-based \cite{radford2021learningtransferablevisualmodels} encoder and latent diffusion models.
%
% After that, SD3 improves its performance by scaling the encoder and decoder architectures.
% %
DALL-E 2 \cite{ramesh2022hierarchical} employs unCLIP while DALL-E 3 \cite{betker2023improving} and PixArt-$\alpha$ \cite{chen2023pixartalphafasttrainingdiffusion} introduce T5 \cite{raffel2023exploringlimitstransferlearning} to enhance alignment.
% %
% Similarly, PixArt-$\alpha$ \cite{chen2023pixartalphafasttrainingdiffusion} utilizes T5 as the text encoder and a more efficient DiT \cite{peebles2023scalablediffusionmodelstransformers} to improve compositional fidelity.
%
More recently, FLUX.1 dev \footnote{\url{https://huggingface.co/black-forest-labs/FLUX.1-dev}} introduces RoPE \cite{su2023roformerenhancedtransformerrotary} to enhance spatial coherence, while FLUX.1 schnell \footnote{\url{https://huggingface.co/black-forest-labs/FLUX.1-schnell}} increases inference speed within 1 to 4 steps.

\subsection{Automatic Prompt Optimization for Image Generation}
T2I models are still facing challenges in text-to-image consistency \cite{wu2023harnessingspatialtemporalattentiondiffusion}, therefore, Automatic Prompt Optimization (APO) \cite{pryzant2023automaticpromptoptimizationgradient}, an automatic technique to optimize the performance of models without training \cite{ramnath2025systematicsurveyautomaticprompt}, has been explored.
Existing APO studies include backpropagation-free optimization method \cite{mañas2024improvingtexttoimageconsistencyautomatic}, Proximal Policy Optimization (PPO)-based \cite{schulman2017proximalpolicyoptimizationalgorithms} reinforcement method \cite{hao2023optimizingpromptstexttoimagegeneration, cao2023beautifulpromptautomaticpromptengineering}, adapters \cite{NEURIPS2024_b2077e6d}, and some products such as MagicPrompt\footnote{\url{https://huggingface.co/Gustavosta/MagicPrompt-Stable-Diffusion}} and PromptPerfect\footnote{\url{https://promptperfect.jina.ai/}}.
However, most existing methods lack error analysis, are limited to specific models, and often rely on coarse-grained evaluators such as CLIPScore \cite{hessel2022clipscorereferencefreeevaluationmetric} or FID \cite{heusel2017gans}, which provide limited reliability in assessing image-text alignment \cite{mañas2024improvingtexttoimageconsistencyautomatic}.
% However, most existing methods lack error analysis, are limited to specific models (e.g., Stable Diffusion \cite{rombach2022highresolutionimagesynthesislatent}), and often rely on coarse-grained evaluators such as CLIPScore \cite{hessel2022clipscorereferencefreeevaluationmetric} or FID \cite{heusel2017gans}, which provide limited reliability in assessing image-text alignment \cite{mañas2024improvingtexttoimageconsistencyautomatic}.

\subsection{Test-Time Scaling for Image Generation}
In recent years, test-time scaling has been extensively studied in large language models \cite{zhao2025surveylargelanguagemodels} with multiple inference samples and a selection mechanism to find the suitable result \cite{lightman2023letsverifystepstep}.
The study \cite{ma2025inferencetimescalingdiffusionmodels} formulates the task as a search problem in noise space and selects the best in $N$ samples.
SANA-1.5 \cite{xie2025sana15efficientscaling} repeats the number of samples rather than denoising steps to scale up the performance.
Also, FK STEERING \cite{singhal2025generalframeworkinferencetimescaling} 
employs FK-IPS \cite{moral2004feynman} to guide the sample path with the high reward.
However, different from those methods operating in the noise space with a fixed input, we formulate the scaling into the input space, which we call ``test-time prompt optimization'' to generate $N$ samples and cluster them to find the optimal one.

\section{Method}
\begin{figure*}[ht!]
\begin{center}
\includegraphics[width=\linewidth]{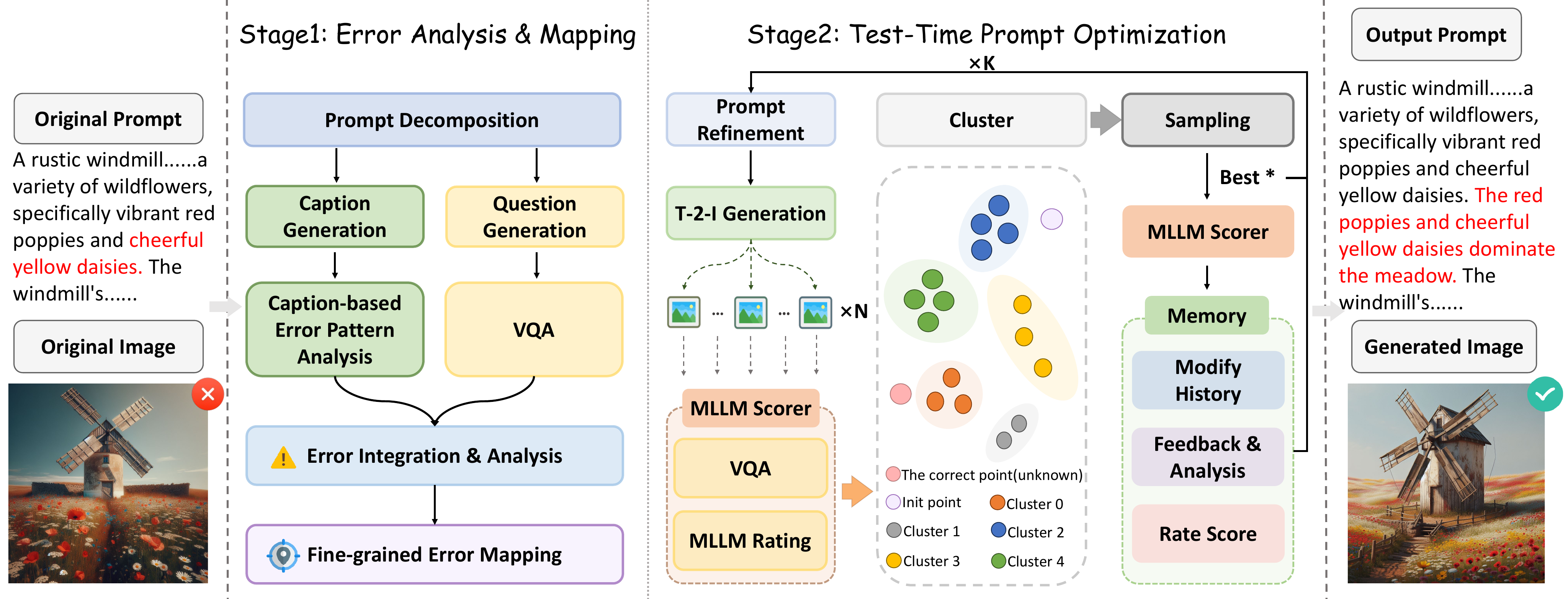}
\end{center}
\caption{
Overview of our proposed multi-agent system for test-time prompt optimization.
GenPilot utilizes a multimodal large language model as the agent.
% %
% The process can be divided into two stages.
%
In stage 1, we first decompose the prompt, then we introduce the error integration strategy based on image caption and VQA results, and map the error to the original prompt.
In stage 2, we introduce the test-time scaling by formulating the problem as a search problem and operating on the input text space.
The test-time prompt optimization is iteratively processed with a refinement agent, an MLLM scorer, a clustering algorithm \cite{macqueen1967some}, and the memory module to sample the optimal currently.
}\label{fig:pipline_3}
\end{figure*}

\subsection{How to Scale at Inference Time for Prompt}
For test-time scaling of textual prompts, we formulate it as a search problem aimed at finding the optimal input for diverse image generation models, which is unknown. 
Unlike the prior work, such as \cite{ma2025inferencetimescalingdiffusionmodels}, which scales the sample noise, our method focuses on the exploration and refinement of the textual inputs.
We operate within a predefined discrete text space, and the prompt is scaled through an iterative process.
GenPilot generates multiple candidate prompts and scores them, then the candidates are clustered to help identify an optimal one.
This optimal candidate then serves as the basis for the subsequent round of optimization.
Consequently, performance is expected to scale positively with the progression of this prompt optimization process.

\subsection{Overall Framework}
As illustrated in Figure \ref{fig:pipline_3}, GenPilot operates in two coarse-grained stages: \textbf{Error Analysis} and \textbf{Test-Time Prompt Optimization}.

Beginning at an initial prompt and image, GenPilot decomposes the prompt into ``meta-sentences'' with an AI agent \cite{Wang_2024}.
Based on these units, GenPilot performs parallel error detection via VQA and captioning, named the error integration strategy.
The VQA-based branch queries object-level details, while the caption-based branch compares captions with the original prompt.
An error-integration agent aggregates the inconsistencies into a comprehensive error list, with another agent mapping each error back to specific prompt segments.
In the test-time prompt optimization stage, a refinement agent generates candidate prompts based on the metadata, including the original prompt and image, and error analysis and mapping.
Detailed definitions and metadata formats are provided in Appendix \ref{MoreDetailedMetadata}.
These candidates are evaluated by an MLLM scorer through VQA and a rating strategy.
GenPilot clusters the prompts and selects the optimal cluster for sampling and image generation.
The memory module is iteratively updated with visual and textual feedback until convergence or a maximum iteration threshold is reached.
The system prompt can be found in Appendix \ref{SystemPromptTemplate}.
\subsection{Error Analysis and Mapping}
\subsubsection{Prompt Decomposition}
Prior work~\cite{wang2024genartistmultimodalllmagent} decomposes prompts into object and background details, but often ignores inter-object relationships, causing semantic errors.
In contrast, we design a coarser-grained prompt decomposition into pieces with an agent that contains objects, relationships, and background information.
For example, given a prompt $P$, the agent segments it as:
\begin{equation}
P = \{ s_1, s_2, \dots, s_n \}
\end{equation}
where $s_n$ denotes sentence pieces.
A more detailed and fine-grained mapping is subsequently performed during the error mapping stage.
% %
% The system prompt can be found in Appendix \ref{SystemPromptTemplate}.
\subsubsection{Error Integration and Localization}
% \subsubsection{Error Analysis and Mapping}
Evaluating text-image alignment by VQA with MLLMs is constrained in complex scenes, leading to unreliable scores.
Therefore, we design an integrated error analysis strategy that combines VQA-based and caption-based detection.
% %
% We utilize two MLLM agents to perform visual question answering (VQA)-based error detection: one for question generation and the other for VQA execution.

\paragraph{Question Generation.}
Inspired by DSG \cite{cho2024davidsonianscenegraphimproving}, we introduce an MLLM agent to generate full coverage questions.
Given a decomposed prompt, the question-generator agent identifies the objects and formulates yes/no questions about object existence, attributes, states, spatial relations, and background information for precise analysis.
% Given a decomposed prompt, the question-generator agent first identifies and breaks down the objects mentioned in the prompt for precise analysis.
% %
% Based on this breakdown, the agent generates yes or no questions covering the existence, the specific attributes, the state, the relative position of a certain object, and so on.

\paragraph{VQA Analysis.}
Each generated question is passed to another MLLM that serves as the VQA agent, who provides a label from $\{\texttt{YES}, \texttt{NO}\}$ and brief explanations to the errors, in the form:
\begin{equation}
e_{vqa_i} = (\text{type}_i,\text{explanation}_i)
\end{equation}
where $\text{type}_i$ is the type of inconsistency and $\text{explanation}_i$ refers to the detailed errors.
The full error set $\mathcal{E}_{vqa}$ is represented as:
% \begin{equation}
% \mathcal{E}_{vqa} = \{ e_1, e_2, \dots, e_n \}
% \end{equation}
\begin{equation}
\mathcal{E}_{vqa} = \{ e_{vqa_1}, e_{vqa_2}, \dots, e_{vqa_n} \}
\end{equation}

\paragraph{Caption-Based Error Analysis.}
For caption-based error analysis, an MLLM generates a detailed caption $C_i$ for image $I_i$, then a comparison agent contrasts $C_i$ with the original prompt $P_i$ to detect semantic discrepancies.
The full error set from caption $\mathcal{E}_{c}$ is represented as:
\begin{equation}
\mathcal{E}_{c} = \{ e_{c_1}, e_{c_2}, \dots, e_{c_n} \}
\end{equation}
where $e_{c_i}$ = $(\text{type}_i,\text{explanation}_i)$ and $e_{c_i}$ denotes the error analyzed from the comparison agent.

\paragraph{Integrated Error Identification.}
In this stage, an MLLM agent functions as an error-integration agent, tasked with synthesizing information from multiple analytical sources, formulated as:
\begin{equation}
\mathcal{E}_{u} = A_{error}(I,P,\mathcal{E}_{c},\mathcal{E}_{vqa})
\end{equation}
where $\mathcal{E}_{u}$ is the finalized error set, $I$ is the original image, and $P$ is the original prompt.
$\mathcal{E}_{c}$ and $\mathcal{E}_{vqa}$ are the error sets from caption- and VQA-based detection.
$A_{error}$ is the error-integration agent.

\paragraph{Error Mapping.}
Error localization maps an identified error to the pieces of the original prompt that lead to it, bridging the abstract error and concrete prompt to support the refinement module.

\subsection{Test-time Prompt Optimization}
% In the test-time prompt optimization stage, we formulate scaling as a search problem over the input prompt, involving a refinement agent for modification, a clustering module to identify the optimum, and a verifier to evaluate prompt quality.  
% The process operates in an iterative loop driven by these key components.

% In the test-time prompt optimization stage, we formulate the scaling task as a search problem for the input prompt by searching and evaluating, containing a prompt refinement agent for modifying, a cluster to find the current optimum, and a verifier to determine what is considered a `good' or high-quality candidate.
% %
% The whole process is designed as an iterative cycle driven by the key components.

\paragraph{Prompt Refinement.}
We first introduce a prompt refinement agent based on metadata to modify the error mapping sentence $m_i \in M$ and generate $N$ diverse candidate modifications $\{m_i^1, m_i^2, \dots, m_i^N\}$, using multiple references to enhance diversity.
Next, each sentence $m_i^j$ is merged into the original prompt $P$ by a branch-merge agent, the process can be formulated as:
\begin{equation}
    P_i^j = A_{merge}(P, m_i^j), \quad j = 1, 2, \dots, N
\end{equation}
where $P_i^j$ denotes the candidate prompts generated by the branch-merge agent $A_{merge}$, which are then passed to T2I model to generate images.

\paragraph{MLLM scorer.}
Subsequently, GenPilot employs an MLLM scorer that acts as a test-time verifier to indirectly evaluate prompt quality via the generated images.  
Inspired by T2I-CompBench \cite{huang2025t2icompbenchenhancedcomprehensivebenchmark}, we design our evaluation rules from the following three aspects: attribute binding including \textit{color}, \textit{number}, \textit{shape}, \textit{state}, and \textit{texture} of the object, relationship and position, and background information and style including the \textit{background description}, the \textit{style}, and \textit{atmosphere}.
A more detailed explanation is provided in Appendix \ref{ADetailedExplanationofScorerSubcategory}.

For each candidate prompt and image pair, the VQA agent analyzes potential inconsistencies based on the question list generated, and a rating agent provides more reliable scores. 
The whole scoring process is defined as:
\begin{equation}
\mathcal{S}(P_i^j) = avg(A_{rate}(I_i^j,P, A_{vqa}(I_i^j, P)))
\end{equation}
where $A_{rate}$ is the rating agent, $A_{vqa}$ denotes the VQA agent, $P_i^j$ is the candidate prompts and $I_i^j$ refers to the corresponding images, and $P$ is the original prompt.

\paragraph{Clustering.}
The scored candidate prompts are then processed with K-Means clustering \cite{macqueen1967some}, including Bayesian updates to progressively identify high-potential prompt candidates.
Initially, each cluster $j$ is assigned the prior probability $P_j = 1/K$, and the candidates are clustered into $K$ groups using K-Means.
Then posterior probabilities $P_j^{\text{post}}$ are computed using the Bayesian update rule, formulated as:
\begin{equation}
\begin{aligned}
P_j^{\text{post}} = \frac{L_j P_j}{\sum_k L_k P_k}
\end{aligned}
\end{equation}
where $L_j$ refers to the likelihood.
The cluster $j^*$ with the highest posterior probability is identified as the best cluster in this round, shown as:
\begin{equation}
\begin{aligned}
j^* = \arg\max_j P_j^{\text{post}}
\end{aligned}
\end{equation}
Following that, a sampled prompt set $s^*$, which contains $m$ candidate sampled prompts from the cluster $j^*$.
The $P_j^{\text{post}}$ serves as the prior distribution for the next round.

\paragraph{Memory.}
For each prompt in the $m$ sample set, we employ the T2I models to generate the image and evaluate them by MLLM scorer.
The average rating and detailed error analysis are stored in the memory module, serving as a historical reference for the subsequent optimization iterations, which can be formulated as:
\begin{equation}
\mathcal{M} \leftarrow \mathcal{M} \cup \{(s^*, I_{s^*}, \mathcal{S}(s^*), A_{sum}(\mathcal{E}_{s^*})\}
\end{equation}
where $s^*$ denotes the sampled prompt set, $I_{s^*}$ refers to the corresponding images and $A_{sum}$ refers to an agent who summarizes the error analyses for $s^*$.

\section{Experiment}
% In this section, we evaluate the effectiveness of our methods with a series of experiments. 
% %
% We mainly conduct the experiments on the GenEval \cite{ghosh2023genevalobjectfocusedframeworkevaluating} benchmark and the DPG-bench \cite{hu2024ellaequipdiffusionmodels} benchmark. 
% %
% The DPG-bench \cite{hu2024ellaequipdiffusionmodels} contains 1,065 lengthy, dense, and complicated textual prompts, while the GenEval \cite{ghosh2023genevalobjectfocusedframeworkevaluating} contains 553 relatively short prompts across six different tasks, including object, counting, and position.

\subsection{Implementation Details}
In the experiment, we employ Qwen2-VL-72B-Instruct \cite{bai2025qwen25vltechnicalreport} as the MLLM agent.
Our method operates with 20 candidate prompts, 5 cluster labels, and undergoes 10 modification cycles. 
Given that users often tend to optimize an image only when initial outputs are unsatisfactory, we construct a challenging subset of 264 prompts selected from the DPG-bench \cite{hu2024ellaequipdiffusionmodels} dataset, with most prompts falling below a threshold of $0.81$, posing significant challenges even for the state-of-the-art models.
Though our method is principally designed for complex and lengthy prompts, we also extended our evaluation to short prompts on the GenEval benchmark \cite{ghosh2023genevalobjectfocusedframeworkevaluating} to ensure a comprehensive assessment of its capabilities.
The results are conducted three times to calculate the average score.
All the system prompts are shown in Appendix \ref{SystemPromptTemplate}.

\subsection{Comparison on DPG-bench subset}
\begin{table*}[!htbp]
  \centering
  % \small 
% \begin{adjustbox}{width=\textwidth,scale=0.7}
  \scalebox{0.8}{
  \begin{tabular}{lcccccc}
    \hline
    \textbf{Model} & \textbf{Average} & \textbf{Global} & \textbf{Entity} & \textbf{Attribute} & \textbf{Relation} & \textbf{Other} \\
    \hline
    DALL-E 3 & 72.04 & \textbf{89.47} & 82.54 & 79.97 & \textbf{90.41} & 63.41 \\
    DALL-E 3 + PE & 72.29 & 85.37 & 82.89 & \textbf{82.98} & 88.88 & \textbf{66.45} \\
    DALL-E 3 + \textbf{Ours} & \textbf{74.08} & \textbf{89.47} & \textbf{83.73} & 81.96 & 88.70 & 60.98 \\
    \hline
    FLUX.1 schnell & 68.16 & 79.12 & 80.33 & 81.02 & 88.24 & \textbf{65.75} \\
    FLUX.1 schnell + PE & 68.38 & 81.32 & 79.69 & 77.54 & 85.99 & 61.64 \\
    FLUX.1 schnell + TTS & 70.26 & \textbf{82.41} & 81.59 & 80.77 & \textbf{90.33} & 64.38 \\
    FLUX.1 schnell + \textbf{Ours} & \textbf{73.32} & 79.12 & \textbf{82.42} & \textbf{83.20} & 89.86 & 61.64 \\
    \hline
    Stable Diffusion v1.4 & 53.16 & 85.71 & 65.23 & 65.70 & 78.63 & 47.37 \\
    % SDv1.4 + Promptist & 43.10 & \textbf{100.00} & 51.48 & 58.01 & 66.41 & 36.84 \\
    Stable Diffusion v1.4 + MagicPrompt & 53.61 & 92.85 & 66.57 & 64.42 & 77.86 & 47.37 \\
    Stable Diffusion v1.4 + BeautifulPrompt & 55.99 & 85.71 & 66.04 & 66.67 & 81.68 & 52.63 \\
    Stable Diffusion v1.4 + PE & 56.08 & 85.71 & 69.27 & \textbf{70.83} & \textbf{88.55} & 47.37 \\
    Stable Diffusion v1.4 + TTS & 55.00 & 71.42 & 66.37 & 66.47 & 77.29 & 41.09 \\
    Stable Diffusion v1.4 + \textbf{Ours} & \textbf{62.12} & \textbf{100.00} & \textbf{71.43} & 67.94 & 79.39 & \textbf{57.89} \\
    \hline
    Stable Diffusion v2.1 & 57.24 & 93.75 & 71.92 & 70.04 & 82.83 & 46.15 \\
    % sd2+promptist & 48.91 & 84.38 & 62.91 & 61.01 & 65.66 & 42.31 \\
    Stable Diffusion v2.1 + MagicPrompt & 58.93 & 93.75 & 70.88 & \textbf{71.81} & 78.79 & 42.31 \\
    Stable Diffusion v2.1 + BeautifulPrompt & 58.04 & 90.63 & 71.58 & 68.94 & 82.32 & 46.15 \\
    Stable Diffusion v2.1 + PE & 56.49 & \textbf{96.88} & 71.23 & 69.60 & \textbf{85.35} & 30.77 \\
    Stable Diffusion v2.1 + \textbf{Ours} & \textbf{61.72} & \textbf{96.88} & \textbf{76.26} & 71.16 & 77.78 & \textbf{53.85} \\
    \hline
    Stable Diffusion 3 & 58.81 & 79.63 & 71.15 & 73.02 & \textbf{84.01} & 51.42 \\
    % SD3 + promptist & 49.83 & 83.33 & 61.67 & 63.65 & 68.03 & 37.14 \\
    Stable Diffusion 3 + MagicPrompt & 59.26 & 83.33 & \textbf{72.82} & 73.02 & 81.41 & 42.86 \\
    Stable Diffusion 3 + BeautifulPrompt & 60.49 & 87.04 & 71.54 & \textbf{73.51} & 80.67 & 48.58 \\
    Stable Diffusion 3 + PE & 58.81 & 81.48 & 70.26 & 70.60 & 82.16 & 31.43 \\
    Stable Diffusion 3 + \textbf{Ours} & \textbf{62.89} & \textbf{88.89} & 72.31 & 68.98 & 79.55 & \textbf{51.43} \\
    \hline
    Sana-1.0 1.6B  & 73.98 & \textbf{85.71} & 83.44 & 81.83 & \textbf{91.63} & 67.12 \\
    Sana-1.0 1.6B + \textbf{Ours} & \textbf{75.38} & 83.78 & \textbf{85.16} & \textbf{83.23} & 90.69 & \textbf{70.97}\\
    \hline
  \end{tabular}
  }
  % \end{adjustbox}
  \caption{Quantitative evaluation of T2I generation performance on DPG-bench challenging dataset comparing GenPilot with generative models and other enhancement methods.
  Our approach consistently achieves superior Average performance and demonstrates notable improvements across various models.
  }
  \label{tab:dpg_performance}
\end{table*}

\subsubsection{Quantitative Evaluation} % number
We evaluate GenPilot on a wide range of T2I models and compare it with the Prompt Engineering method (PE), naive test time scaling methods (TTS), and SD-based methods: MagicPrompt and BeautifulPrompt \cite{cao2023beautifulpromptautomaticpromptengineering}.
%
% The system prompt used in PE is shown in Appendix \ref{SystemPromptTemplate}.
According to Table \ref{tab:dpg_performance}, GenPilot successfully improves the performance in the overall ``Average'' score on all models tested.
For example, the average score improves from $72.04$ to $74.08$ on DALL-E 3, from $68.16$ to $73.32$ on FLUX.1 ($68.38$ by PE and $70.26$ by TTS), from $73.98$ to $75.38$ on Sana-1.0 1.6B, and from $53.16$ to $62.12$ on SDv1.4 ($55.99$ and $53.61$ by BeautifulPrompt and MagicPrompt, respectively).
Similar gains are observed on SDv2.1 and SD3, indicating the robustness and generalizability of GenPilot, revealing its ability to enhance weaker models while refining top-tier ones.
Compared to Sana-1.0 1.6B ($73.98$), GenPilot enables DALL-E 3 ($74.08$) to surpass it and FLUX.1 schnell ($73.32$) to perform comparably, highlighting GenPilot's effectiveness through test-time prompt optimization.
Although some subcategories show slightly lower scores, GenPilot achieves the highest performance on average, revealing a balance in optimization across different aspects.

\subsubsection{Qualitative Evaluation} % image
\begin{figure*}[!htbp]
\begin{center}
\includegraphics[width=0.98\linewidth]{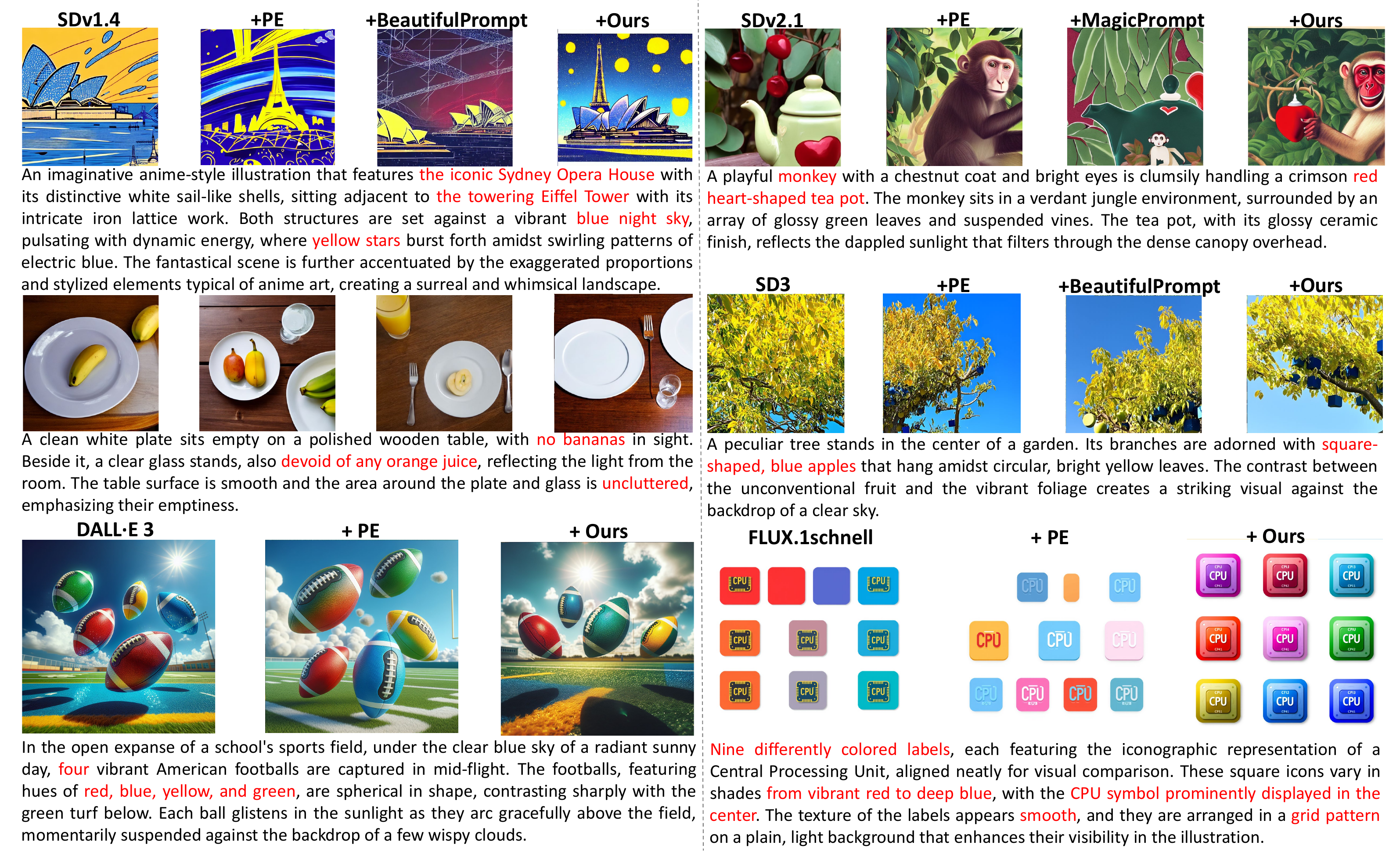}
\end{center}
\caption{
Qualitative comparison with different methods on the DPG-bench challenging dataset on different generative models.
The left columns display two generations from SDv1.4 and one from DALL-E 3. 
The right columns present the results from SDv2.1, SD3, and FLUX.1 schnell.
For the SD series, we select the best from BeautifulPrompt and MagicPrompt, along with the PE methods for comparison.
GenPilot consistently generates error-free images across all scenarios, demonstrating its superiority in synthesizing high-quality and accurate images.
}\label{fig:show_dpg_sd_2}
\end{figure*}
On the second row on the left in Figure \ref{fig:show_dpg_sd_2}, the image generated by SDv1.4 with GenPilot effectively excludes the unwanted items, in contrast to the other three images, which fail this exclusion and contain them to varying extents.
These qualitative examples in Figure \ref{fig:show_dpg_sd_2} vividly illustrate the effectiveness and generalization ability of GenPilot in handling challenging prompts, including accurate attribute binding such as counting, complex compositions, spatial reasoning, unrealistic description and the effective processing of negative constraints.
More qualitative analysis can be found at Appendix \ref{MoreResultsonDPG-bench}.

\subsection{Comparison on GenEval benchmark}
% In this section, we conduct a comprehensive experiment on the GenEval benchmark with a relatively short prompt.
\subsubsection{Quantitative Evaluation} % number
\begin{table*}[!htbp]
  \centering
  \scalebox{0.7}{
  \begin{tabular}{lccccccc}
    \hline
    \textbf{Model} & \textbf{Overall} & \textbf{Position} & \textbf{Color\_Attr} & \textbf{Colors} & \textbf{Sin\_Obj} & \textbf{Two\_Obj} & \textbf{Counting} \\
    \hline
    FLUX.1 schnell & 65.82 & 29.00 & 44.50 & 
 76.06 & \textbf{99.69} & \textbf{86.62} & 59.06    \\
    FLUX.1 schnell + PE & 66.59 & 31.75 & 46.50 & 80.32 & 99.06 & 85.35 & 56.56    \\
    FLUX.1 schnell \textbf{+ Ours} & \textbf{69.60} & \textbf{41.50} & \textbf{52.25} & \textbf{81.38} & 97.19 & 84.60 & \textbf{60.62}    \\
    \hline
    PixArt-$\alpha$ & 46.73 & 8.25 & 7.00 & 77.66 & \textbf{98.44} & \textbf{50.00} & 39.06 \\
    PixArt-$\alpha$ + PE & 45.98 & 8.50 & 8.50 & 71.54 & 97.81 & 45.45 & 44.06 \\ % Updated data for pixart+ICL
    PixArt-$\alpha$ \textbf{+ Ours } & \textbf{48.54} & \textbf{9.25} & \textbf{9.25} & \textbf{81.91} & 95.31 & 49.24 & \textbf{46.25} \\
    \hline
  \end{tabular}
  }
  \caption{Quantitative results on GenEval benchmark.
  All scores are reported as percentages (\%). The `\%' symbol is omitted for brevity.
  Sin\_Obj refers to a single object, and Two\_Obj represents two objects.
  Color\_attr is the color attribute in short.
  GenPilot demonstrates superior overall generation ability both on FLUX.1 schnell and PixArt-$\alpha$, with a great improvement on most of the subcategories.
  }
  \label{tab:geneval_performance}
\end{table*}

\subsubsection{Qualitative Evaluation} % image
\begin{figure*}[!htbp]
\begin{center}
\includegraphics[width=0.8\linewidth]{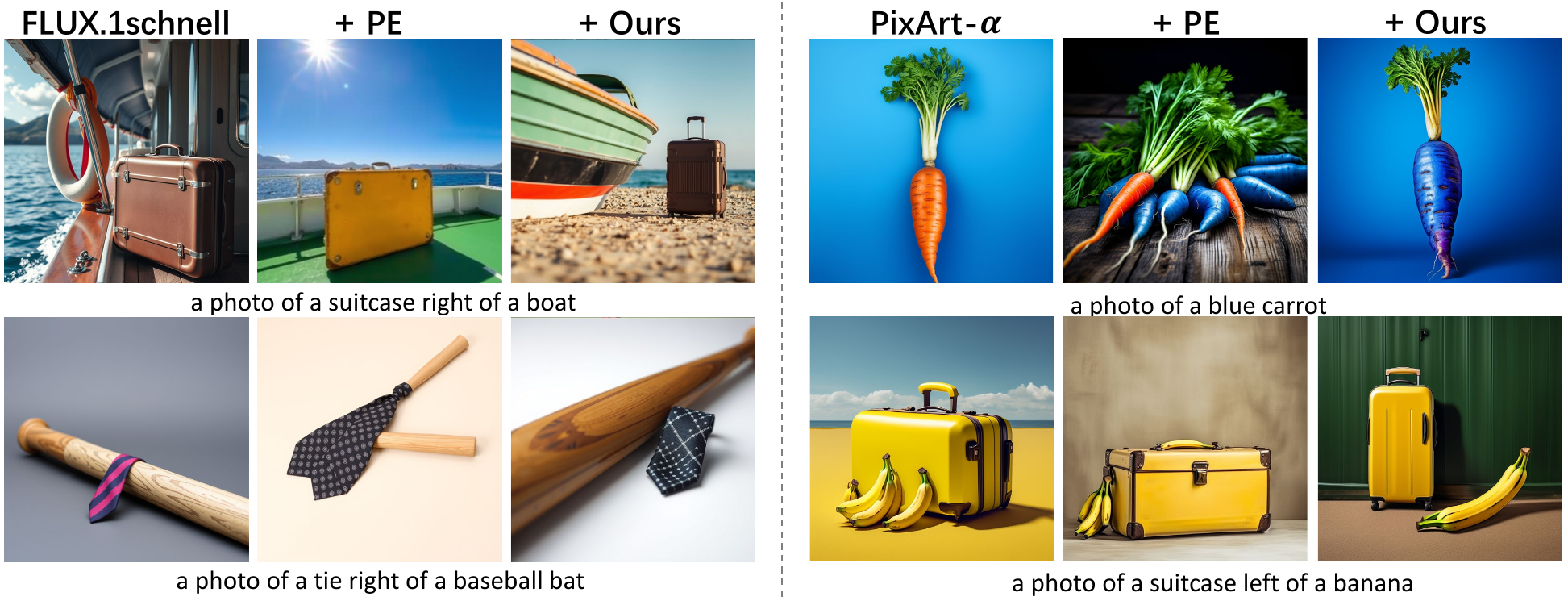}
\end{center}
\caption{
Qualitative examples on GenEval.
The left columns show the comparison of FLUX.1 schnell, FLUX.1 schnell and PE for enhancement, and FLUX.1 schnell with GenPilot.
The right columns provide the results of PixArt-$\alpha$, PixArt-$\alpha$ and PE for enhancement, and PixArt-$\alpha$ with GenPilot.
GenPilot achieves great success in both position processing and unrealistic prompt generation, highlighting its potential and generalization to improve the quality of images.
}\label{fig:geneval_result}
\end{figure*}

\begin{table*}[ht]
  \centering
  \scalebox{0.8}{
  \begin{tabular}{lccccccc}
    \hline
    \textbf{Model} & \textbf{Overall} & \textbf{Position} & \textbf{Color\_Attr} & \textbf{Colors} & \textbf{Sin\_Obj} & \textbf{Two\_Obj} & \textbf{Counting} \\
    \hline
FLUX.1 schnell  & 65.82 & 29    & 44.5  & 76.06 & \textbf{99.69} & \textbf{86.62} & 59.06 \\
+ Ours-M    & 66.05 & 35.75 & 46.75 & 74.73 & 98.44 & 82.83 & 57.81 \\
+ Ours-C    & 66.27 & 35.75 & 46.75 & 77.66 & 97.19 & 83.08 & 57.19 \\
\textbf{+ Ours}     & \textbf{69.60} & \textbf{41.50} & \textbf{52.25} & \textbf{81.38} & 97.19 & 84.60 & \textbf{60.62} \\
    \hline
  \end{tabular}
  }
  \caption{Ablation study results on different variants of our method on GenEval with `\%' omitted.
  ``+ Ours-M'' refers to FLUX.1 schnell with GenPilot but removing the memory module, and ``+ Ours-C'' represents the variant without clustering. 
  GenPilot performs the best with comprehensive improvements, illustrating the effectiveness of these modules.
  }
  \label{tab:geneval_ablation}
\end{table*}

\begin{table*}[t]
\centering
\scalebox{0.8}{
\begin{tabular}{lcccccc}
\hline
    \textbf{Model} & \textbf{Average} & \textbf{Global} & \textbf{Entity} & \textbf{Attribute} & \textbf{Relation} & \textbf{Other} \\
\hline
FLUX.1 schnell & 68.16 & 79.12 & 80.33 & 81.02 & 88.24 & 65.75 \\
+ MiniCPM-V 2.0 & 69.82 & 76.92 & 82.32 & 82.01 & 84.00 & 76.92 \\
+ Qwen2.5-VL-72B & 73.32 & 79.12 & 82.42 & 83.20 & 89.86 & 61.64 \\
\hline
% + BLIP-2 (Captioner) & 69.22 & -- & -- & -- & -- & -- \\
% \hline
\end{tabular}
}
\caption{Ablation study on different MLLM agents and captioners in GenPilot. 
MiniCPM-V 2.0 achieves competitive results compared to Qwen2.5-VL-72B.}
\label{tab:mlm-ablation}
\end{table*}

As illustrated in Table \ref{tab:geneval_performance}, GenPilot applied to the two base models on GenEval, including FLUX.1 schnell and PixArt-$\alpha$, compared to the Prompt Engineering(PE).
GenPilot improves FLUX.1 schnell from $65.82\%$ to $69.60\%$, outperforming PE ($66.59\%$) with notable gains in position, color, and number-related tasks.
Similarly, PixArt-$\alpha$ with GenPilot achieves $48.54\%$, surpassing both the base model ($46.73\%$) and its PE-enhanced version ($45.98\%$).
These results highlight the capability of GenPilot to improve the image quality and text-to-image consistency across models and prompt types.
However, in subcategories such as single- and dual-object scenes, where the base models are already highly proficient, GenPilot shows comparable or slightly lower performance, aligning with its design goal of refining unsatisfactory generations.

\begin{table*}[t]
\centering
\scalebox{0.8}{
\begin{tabular}{cccccc}
\hline
Iter & Cand. & Clust. & AvgTime (s) & GenRatio (\%) & AvgOptTime (s) \\
\hline
1 & 1 & 1 & 29.0 & 52.4 & 13.8 \\
3 & 1 & 1 & 100.0 & 30.4 & 69.6 \\
5 & 1 & 1 & 117.4 & 41.6 & 68.6 \\
% 7 & 1 & 1 & 157.6 & 30.5 & 109.6 \\
7 & 3 & 3 & 128.4 & 38.0 & 79.6 \\
% 7 & 5 & 3 & 264.8 & 26.0 & 196.0 \\
% 7 & 5 & 5 & 266.6 & 27.3 & 193.8 \\
% 7 & 7 & 5 & 275.2 & 21.2 & 216.8 \\
% 7 & 15 & 7 & 380.4 & 18.1 & 311.6 \\
\hline
\end{tabular}
}
\caption{Latency analysis of GenPilot under different configurations. Iter: iteration number; Cand.: candidate prompts; Clust.: number of clusters. AvgTime includes both T2I generation and optimization time, while AvgOptTime isolates optimization overhead.}
\label{tab:latency-config}
\end{table*}
\begin{table}[t]
\centering
\scalebox{0.8}{
\begin{tabular}{lc}
\hline
\textbf{Model} & \textbf{Average} \\
\hline
FLUX.1 schnell & 68.16 \\
+ BLIP-2 (Captioner) & 69.22 \\
+ Qwen2.5-VL-72B & 73.32 \\
\hline
\end{tabular}
}
\caption{Ablation study on different captioner modules in \textsc{GenPilot}. 
Replacing the captioner with BLIP-2 improves over the baseline but remains below Qwen.}
\label{tab:captioner-ablation}
\end{table}

\begin{table}[htbp]
  \centering
  \scalebox{0.8}{
  \begin{tabular}{ccc}
    \hline
 \textbf{VQA-based} & \textbf{Caption-based} & \textbf{Integration}\\
    \hline
3.78 & 3.95   & \textbf{4.62}  \\

    \hline
  \end{tabular}
  }
  \caption{
  Comparison on the accuracy and coverage of error analysis rated by GPT-4o on VQA-based methods, caption-based method, and the integration, highlighting the importance of components in GenPilot.}
  \label{tab:analysis}
\end{table}

Figure \ref{fig:geneval_result} shows the qualitative results of FLUX.1 schnell and PixArt-$\alpha$ on the GenEval benchmark.
As shown in Figure \ref{fig:geneval_result}, with GenPilot, FLUX.1 schnell and PixArt-$\alpha$ can accurately generate the position-related image and unrealistic prompt, compared to failures in PE and base models.
%
% \begin{table*}[ht]
%   \centering
%   \scalebox{0.8}{
%   \begin{tabular}{lccccccc}
%     \hline
%     \textbf{Model} & \textbf{Overall} & \textbf{Position} & \textbf{Color\_Attr} & \textbf{Colors} & \textbf{Sin\_Obj} & \textbf{Two\_Obj} & \textbf{Counting} \\
%     \hline
% FLUX.1 schnell  & 65.82 & 29    & 44.5  & 76.06 & \textbf{99.69} & \textbf{86.62} & 59.06 \\
% + Ours-M    & 66.05 & 35.75 & 46.75 & 74.73 & 98.44 & 82.83 & 57.81 \\
% + Ours-C    & 66.27 & 35.75 & 46.75 & 77.66 & 97.19 & 83.08 & 57.19 \\
% \textbf{+ Ours}     & \textbf{69.60} & \textbf{41.50} & \textbf{52.25} & \textbf{81.38} & 97.19 & 84.60 & \textbf{60.62} \\
%     \hline
%   \end{tabular}
%   }
%   \caption{Ablation study results on different variants of our method on GenEval with `\%' omitted.
%   %
%   ``+ Ours-M'' refers to FLUX.1 schnell with GenPilot but removing the memory module, and ``+ Ours-C'' represents the variant without clustering. 
%   %
%   GenPilot performs the best with comprehensive improvements, illustrating the effectiveness of these modules.
%   }
%   \label{tab:geneval_ablation}
% \end{table*}
The qualitative results reveal the potential of the generalization ability and effectiveness of GenPilot to improve the text-to-image alignment.
More qualitative results are in Appendix \ref{MoreResultsonGenEval}.

\subsection{Ablation Study}

% \begin{table*}[ht]
%   \centering
%   \scalebox{0.7}{
%   \begin{tabular}{lccccccc}
%     \hline
%     \textbf{Model} & \textbf{Overall} & \textbf{Position} & \textbf{Color\_Attr} & \textbf{Colors} & \textbf{Sin\_Obj} & \textbf{Two\_Obj} & \textbf{Counting} \\
%     \hline
% FLUX.1 schnell  & 65.82 & 29    & 44.5  & 76.06 & \textbf{99.69} & \textbf{86.62} & 59.06 \\
% + Ours-M    & 66.05 & 35.75 & 46.75 & 74.73 & 98.44 & 82.83 & 57.81 \\
% + Ours-C    & 66.27 & 35.75 & 46.75 & 77.66 & 97.19 & 83.08 & 57.19 \\
% \textbf{+ Ours}     & \textbf{69.60} & \textbf{41.50} & \textbf{52.25} & \textbf{81.38} & 97.19 & 84.60 & \textbf{60.62} \\
%     \hline
%   \end{tabular}
%   }
%   \caption{Ablation study results on different variants of our method on GenEval with `\%' omitted.
%   %
%   ``+ Ours-M'' refers to FLUX.1 schnell with GenPilot but removing the memory module, and ``+ Ours-C'' represents the variant without clustering. 
%   %
%   GenPilot performs the best with comprehensive improvements, illustrating the effectiveness of these modules.
%   }
%   \label{tab:geneval_ablation}
% \end{table*}
To comprehensively evaluate the contributions of each core component in GenPilot, we conduct ablation studies on the GenEval benchmark, using FLUX.1 schnell.
In this section, we systematically evaluate the impact of the error integration, the clustering, and the memory module.
As shown in Table \ref{tab:geneval_ablation}, GenPilot achieves the highest score of $69.60\%$, and the score without memory is $66.05\%$, and the score without clustering is $66.27\%$, declining across various subcategories.
The results demonstrate the significance of the memory module and clustering algorithm, as the memory provides references and clustering optimizes the search space on text, iteratively scaling up the performance of optimization.
Simultaneously, even removing those key components, GenPilot variants still outperform the base model, revealing the effectiveness of the rest modules in GenPilot.

We further study the effect of different MLLM agents and captioners in GenPilot.
As shown in Table~\ref{tab:mlm-ablation}, replacing Qwen2.5-VL-72B-Instruct with MiniCPM-V 2.0 \cite{yao2024minicpm} yields slightly lower performance but still outperforms FLUX.1 schnell, demonstrating the flexibility of GenPilot across different MLLM backbones.
Moreover, as shown in Table~\ref{tab:captioner-ablation}, replacing the captioning module with BLIP-2 \cite{li2023blip} also improves over the baseline, though its relatively simpler captions result in lower gains compared to Qwen.
These results highlight the modularity of GenPilot in adapting to different components.

We further investigate the latency of GenPilot. 
Table~\ref{tab:latency-config} reports the average time under different configurations.
During the inference stage, the time cost of optimization increases based on the number of iterations, error, candidate prompt, sentence, the T2I models, and the image batch size.
Moreover, we employ parallelization and early stopping strategies to alleviate the time cost in practice. 

Meanwhile, we explore the performance of the error integration strategy by GPT-4o \cite{openai2024gpt4technicalreport} to score the quality of error analysis in VQA-based, caption-based, and integration results from 1 to 5, and 5 is regarded as the best.
As illustrated in Table \ref{tab:analysis}, though analysis from both methods provides effective information, the integration strategy highlights the effectiveness of full coverage and accuracy with a $4.62$ score.
A qualitative comparison example can be found at Appendix \ref{errorcase}.

% On average, GenPilot costs $8$ minutes for one prompt, including the time cost of the image generation process of SDv1.4, when iteration is 3, cluster number is 3, the total number of candidate prompts is 7, and an image batch contains 3 images.
% Moreover, we employ parallelization and early stopping strategies to alleviate the time cost in practice. 

% We further investigate the latency of GenPilot. 
% Table~\ref{tab:latency-config} reports the average time under different configurations.
% %
% The overall latency grows with more iterations, candidate prompts, and clusters, which reflects the trade-off between optimization depth and efficiency.
% %
% The optimization overhead (AvgOptTime) accounts for a large proportion of the cost, as it involves iterative prompt generation, image creation, and MLLM-based evaluation. 
% %
% Table~\ref{tab:latency-compare} compares GenPilot with other prompt enhancement methods.
% %
% While our advanced multi-iteration setting (\textsc{Ours-7}) is slower, the minimal configuration (\textsc{Ours-1}) is competitive, even faster than MagicPrompt, demonstrating the flexible latency-quality balance.
%
% Moreover, we employ parallelization and early stopping strategies to alleviate the time cost in practice. 

% On average, GenPilot costs $8$ minutes for one prompt, including the time cost of the image generation process of SDv1.4, when iteration is 3, cluster number is 3, the total number of candidate prompts is 7, and an image batch contains 3 images.

More experiments on visualization of clustering are provided on Appendix \ref{ClusteringAnalysis}, semantic analysis on embedding is at Appendix \ref{SemanticAnalysisonEmbedding}, and analysis on POS distribution shift is shown in Appendix \ref{AnalysisonPOSDistributionShift}.
% and the patterns on error analysis and refinement can be found at Appendix \ref{DetailedPatternonErrorandOptimizationAnalysis}.

\subsection{Patterns on Error Analysis and Refinement}
We release 35 patterns and their corresponding refinement strategy summarized by GPT-4o, along with cases for better understanding in Appendix \ref{DetailedPatternonErrorandOptimizationAnalysis}.

\section{Conclusion}
In this work, we propose GenPilot, a flexible and effective test-time prompt optimization multi-agent system for enhanced text-to-image generation, aiming to address the semantic gap and the compositionality catastrophe, especially for complicated and lengthy prompts.
Unlike previous approaches, GenPilot performs test-time scaling directly on the input prompt space, formulating it as a search problem to find the optimal prompts for T2I models, iteratively refining the prompt with clustering algorithm.
The system integrates modular agents for error analysis, prompt editing, multi-modal LLM scoring, and memory-based feedback to support dynamic adjustment.
Extensive experiments on GenEval and DPG-bench demonstrate the effectiveness and superiority of GenPilot over other methods, highlighting the potential of test-time prompt optimization for enhancing T2I generation.
We further release a set of common error patterns and refinement strategies, providing a practical resource for future research on prompt controllability and optimization.

\section*{Limitations}
Despite the improved performance of GenPilot in various scenarios, there are still a few challenges to address. 
Firstly, although our framework avoids model fine-tuning, it introduces additional computation time during inference, which may be non-trivial in latency-sensitive applications.
Meanwhile, the performance of GenPilot is influenced by the multimodal large language models used for the agent, which may harm the performance if users utilize a less capable MLLM that lacks sufficient understanding of multimodal information.

% \section*{Acknowledgments}

\section*{Ethics Statement}
In this work, we utilize Qwen2-VL 72B Instruct and GPT-4o as tools for agent or evaluation, along with the dpg-bench and GenEval datasets. 
We fully considered the ethical problems when applying the large language models.
The DPG-bench dataset is licensed under Apache 2.0, and the GenEval dataset is available under the MIT license.
Our usage strictly follows the licenses and their intended purposes.
The data we utilize do not contain any information about unique individual people.

\section*{Acknowledgements}
This work is jointly sponsored by National Natural Science Foundation of China (62141608, 62236010, 62576339), and Beijing Natural Science Foundation (L252033).

\bibliography{main}

\appendix
\section{A Detailed Explanation of Metadata}
\label{MoreDetailedMetadata}
To offer more structural data for the agent to better understand, we design a structural data called metadata.
Initially, the main components in metadata are error analysis, error mapping, question list, history feedback, the original prompt, and the original image generated from that prompt.
We provide the error analysis and mapping, along with the original prompt, image, and history for prompt refinement, and offer the question list for the MLLM scorer.
With the structured metadata, the agent is capable of better understanding the context and efficiently retrieving data.

\section{A Detailed Explanation of Scorer Subcategory}
\label{ADetailedExplanationofScorerSubcategory}
We design the rules from the following three aspects inspired by T2I-CompBench \cite{huang2025t2icompbenchenhancedcomprehensivebenchmark}.

\textbf{Attribute binding:} Attribute binding refers to the ability to correctly associate specific properties with the object as described in the prompt, including color, number, shape, state, and texture of the object.
\begin{itemize}
    \item The color is used to evaluate whether the correct color is applied to a certain object or not, especially when multiple objects have different color specifications.
    \item The number specifies the exact count of objects.
Models might struggle with precise counts, failing to make the very approximate number of different objects.
\item 
The shape refers to the external form or geometric shape of an object, ranging from simple and concrete forms to complex and abstract structures.
For example, in the prompt ``A person with a muscular build'', muscular build refers to the shape of the human.
\item 
The state is a broad category referring to the condition, mode of being, phase, or dynamic activity of an object or entity at a particular time.
It contains physical conditions for instance, ``ripe'' in ``ripe bananas'', the action, such as the ``running'' in the prompt ``A dog running in a field'', the emotional state, for example, the ``surprised'' in ``A surprised cat'', and the functional state such as ``open'' in ``An open door'', and the texture describes the surface of the object, including smoothness, roughness, softness and so on.
\end{itemize}

\textbf{Relationship and position:} In addition to the attribute of the object, prompts often include information about how these objects are interconnected and their positions within the scene.
These relationships involve various types of interactions.
For example, one object acting upon another, such as ``a dog catching a ball'', and the objects in occlusion, such as ``a tree partially obscuring a view of the house'', and simple containment or support, such as ``Apples in a basket''.
Similarly, positional information describes where the object is located, either relative to one another or at the absolute position within the image frame.

\textbf{Background information and style:} Finally, we also defined a further descriptive aspect, the background information and style.
The background information encompasses details about the scene that are distinct from the main object, including the style and overall atmosphere in the image.
\section{Clustering Analysis}
\label{ClusteringAnalysis}

\begin{figure}[ht]
    \centering
    \includegraphics[width=\linewidth]{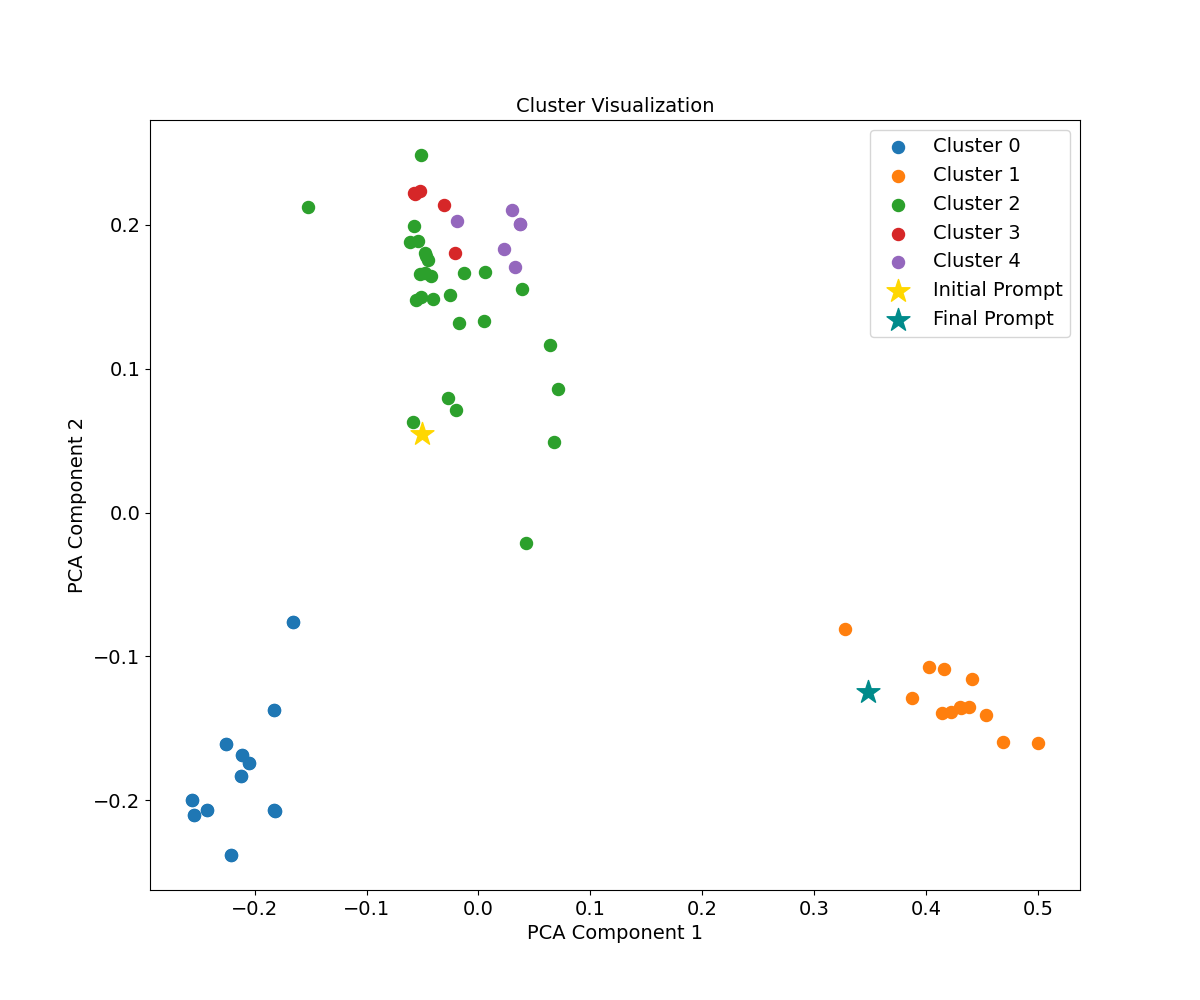}
    \caption{Visualization of clustering result on one case with the number of clusters set to be 5.}
    \label{fig:cluster}
\end{figure}

Figure \ref{fig:cluster} vividly shows how our clustering algorithm works.
The initial prompt (the yellow star) is close to cluster 2 in green, next to cluster 3 in red and cluster 4 in purple.
However, cluster 0 in blue and cluster 1 in orange are far from the initial point.
The relevant score of cluster 1 is 5.0 on average, which indicates it as the best prompt this turn, while clusters 2, 3, and 4 with a lower score, such as 4.3 on average.
Initially, the candidate prompts generated from the prompt refinement agent might still predominantly cluster around.
As iterations progress, GenPilot explores more directions, including the clusters 0 and 1 illustrated in Figure \ref{fig:cluster}.
In this case, cluster 1 represents the optimized area that T2I models prefer to generate high-quality images.
By generating multiple samples and scoring them into clusters, GenPilot successfully scales the prompts and optimizes them, revealing the effectiveness and potential of the test-time prompt optimization for improving the image quality.

Another example with an image and a prompt can be found at Appendix \ref{MoreDetailedCaseAnalysis}.

% \subsubsection{Error Case Analysis}

\section{Semantic Analysis on Embedding}
\label{SemanticAnalysisonEmbedding}
\begin{table}[ht]
  \centering
  \begin{tabular}{lcc}
    \hline
    \textbf{Method} & \textbf{GenEval} & \textbf{DPG-bench}  \\
    \hline
    Origin       & 0.2443 & 0.3705  \\
    BeautifulPrompt & 0.2573 & 0.3527  \\
    PE       & \textbf{0.3193} & 0.3724 \\
    \textbf{Ours}      & 0.2981 & \textbf{0.3944}  \\
    \hline
  \end{tabular}
  \caption{Comparison of the semantic similarity analysis with extremely detailed descriptions by GPT-4o.}
  \label{tab:consine_prompts}
\end{table}

\begin{figure*}[ht]
    \centering
    \includegraphics[width=\linewidth]{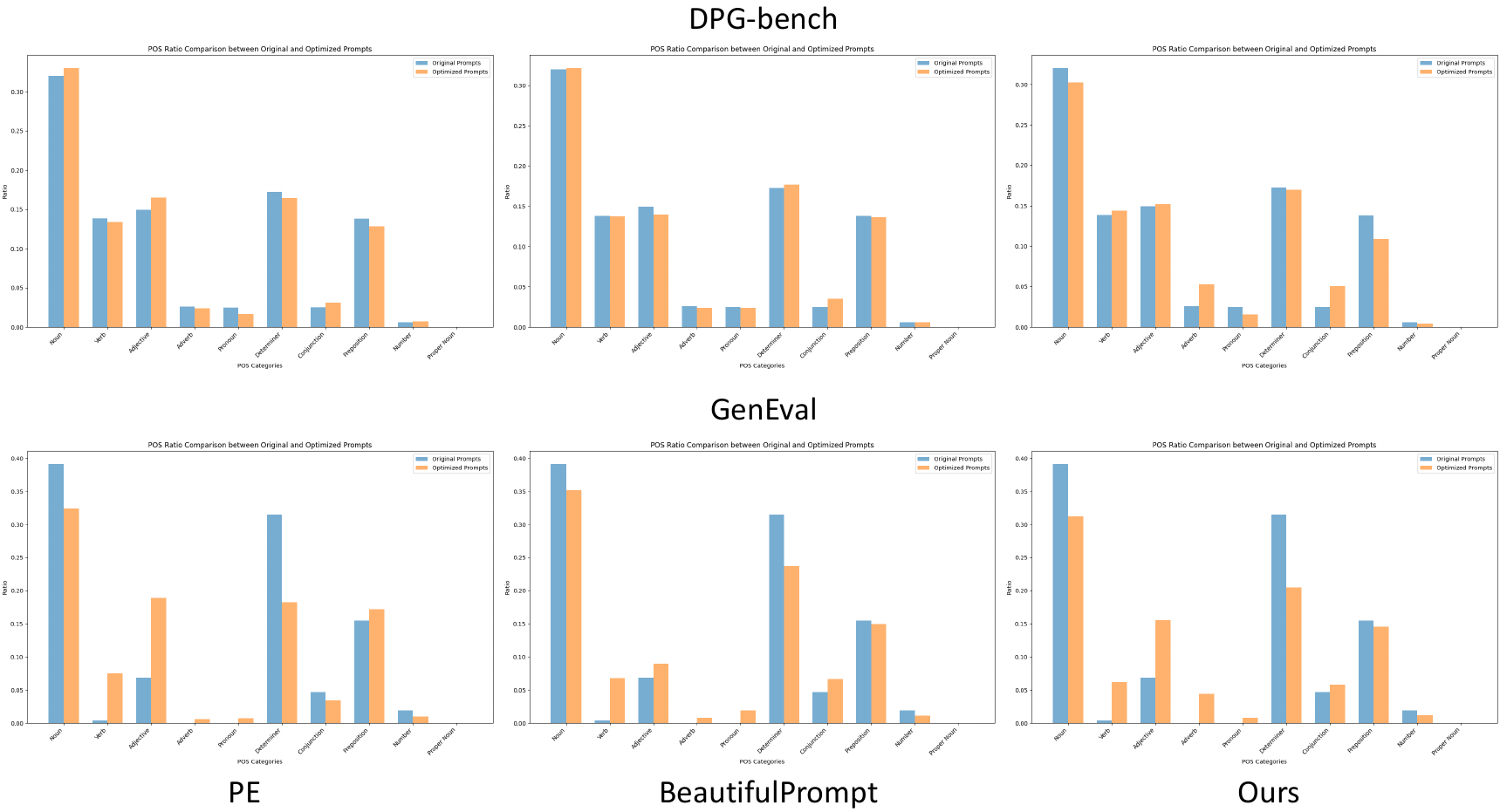}
    \caption{Analysis on POS Distribution Shift of PE, BeautifulPrompt, and Ours compared to original prompt on DPG-bench challenging subset and GenEval benchmark.}
    \label{fig:ratio}
\end{figure*}

We conduct a semantic similarity analysis at the embedding level to evaluate whether prompt optimization leads to richer textual descriptions on the GenEval and DPG benchmark.
We generate extremely detailed and specific descriptions as reference prompts using GPT-4o with the particular instruction shown in Appendix \ref{SystemPromptTemplate}.
We then measure the cosine similarity between each method’s prompt embeddings and the reference for completeness and semantic content.
As shown in Table \ref{tab:consine_prompts}, our method achieves the highest average similarity scores on DPG-bench, given the highest score of performance on DPG-bench in Table \ref{tab:dpg_performance}, indicating that GenPilot introduces meaningful and effective details into the original prompt.
And GenPilot is highly competitive on GenEval (0.2981) against other methods like Origin, BeautifulPrompt, and PE.
When the prompt is relatively short and simple, rewriting or expanding the abstract prompts significantly improves semantic richness, which positively influences generation.
However, on GenEval, we observe that though PE reaches the highest score of similarity, the whole performance of PE is lower than ours when the generative model is FLUX.1 schnell, and even lower than PixArt-$\alpha$ itself.
Therefore, higher semantic similarity for more information included does not always lead to better visual results.
Simply expanding the prompt, especially for complex and lengthy prompts, may not enhance the image result obviously.
In contrast, GenPilot consistently turns the semantic gains into meaningful performance improvements, highlighting its effectiveness and necessity.

% \subsubsection{Analysis on POS Distribution Shift}
\section{Analysis on POS Distribution Shift}
\label{AnalysisonPOSDistributionShift}
To explore the impact of the linguistic structure of generated prompts, we conduct a part-of-speech (POS) level analysis comparing the original prompts and the optimized ones with NLTK \cite{bird2006nltk}.
All tools and functions were used with default settings.
We focus on adjectives, nouns, verbs, adverbs, pronouns, and so on.
A common trend can be found among PE, BeautifulPrompt, and Ours, revealing that adding adjectives may help with more specific information for image generation.
Both on DPG-bench and GenEval, our method increases the proportion of adjectives and proper nouns, indicating that prompts generated by GenPilot tend to be more descriptive via adjectives and more specific via proper nouns.
%
% The detailed information can be found at Appendix \ref{}.

% \subsubsection{Visualization Analysis on PCA}
% \begin{figure}[htbp]
%     \centering
%     \includegraphics[width=\linewidth]{img/ours_combination.png}
%     \caption{a}
%     \label{fig:ours_combination}
% \end{figure}

\section{More Detailed Case Analysis}
\label{MoreDetailedCaseAnalysis}

\begin{figure}[ht]
    \centering
    \includegraphics[width=\linewidth]{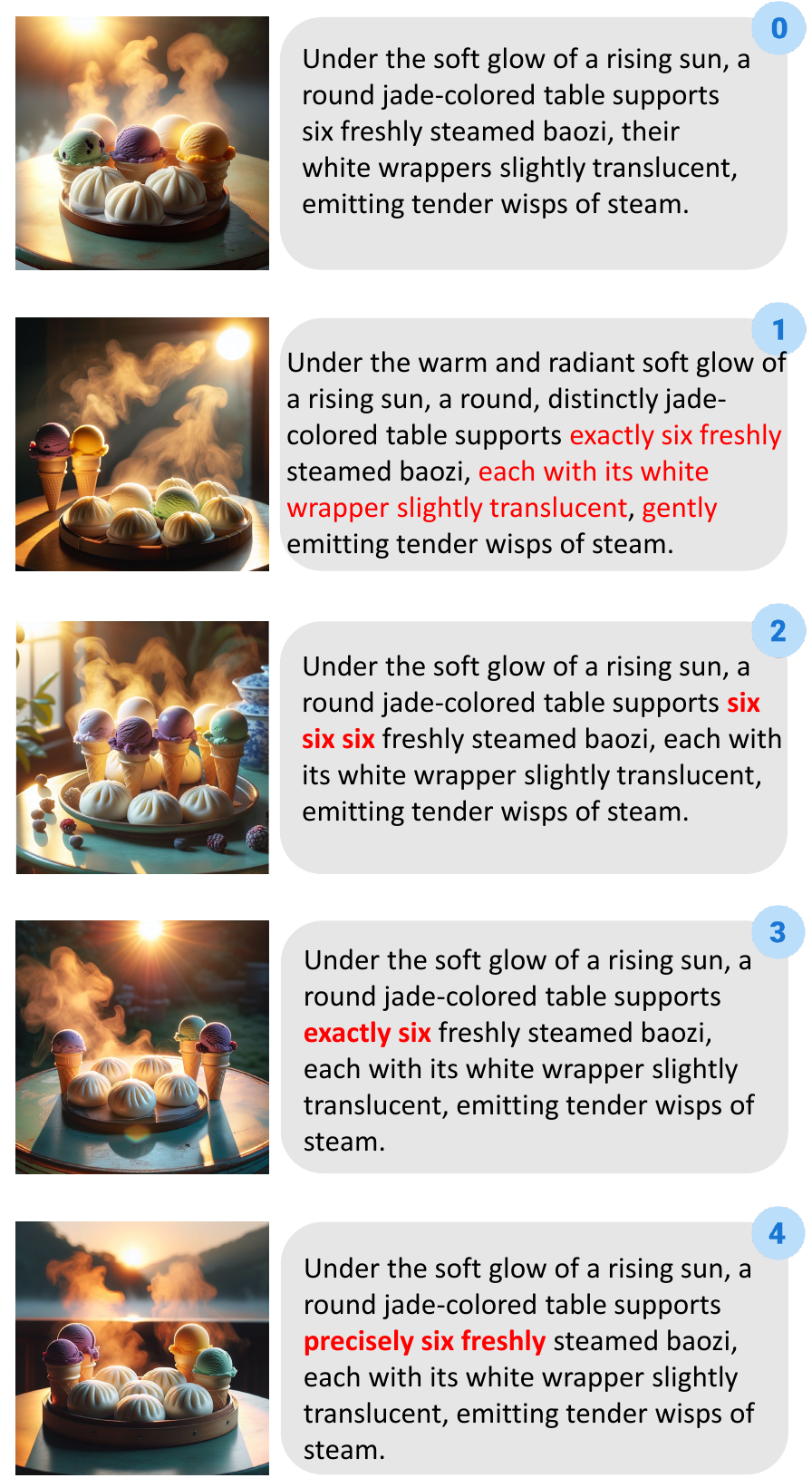}
    \caption{A detailed sample of iterations and the results.
    0 represents the initial start point.
    GenPilot optimizes the sentence with the error ``the number of baozi'' and achieves the accurate synthesis on the fourth round.}
    \label{fig:cluster_case_2}
\end{figure}

In this section, we provide a more detailed case during the iterations, as shown in Figure \ref{fig:cluster_case_2}.
The main error, according to the error mapping sentence in the original picture in the first row, is the number of baozi.
In the original prompt, baozi should be 6, while in the image, it only has three.
The best sampled prompt in the next round modified the prompt with ``exactly'' and some other specific descriptions, rated 4.1 in the end.
In the second round, the prompt optimization agent tries to emphasize the number by repeating the keyword of six.
However, it remains 5 baozi in the image, rated 4.3 by the MLLM scorer.
Next round, the prompt optimization agent concludes the failures of the previous round, and makes an attempt to emphasize by adding an adverb.
In round three, an image with 5 clearly visible baozi is generated, which is a minor improvement compared to the earlier round.
For round 4, prompt optimization tries to change the adverb, which turn out to be successful, rated 5 in the end.
After that, the correct modification, the image, and candidate prompt will be stored, as a stop signal for this error.

% \section{Example Appendix}
% \label{sec:appendix}
\section{System Prompt Template}
\label{SystemPromptTemplate}
\begin{figure}[htbp]
    \centering
      \scalebox{0.9}{
    \includegraphics[width=\linewidth]{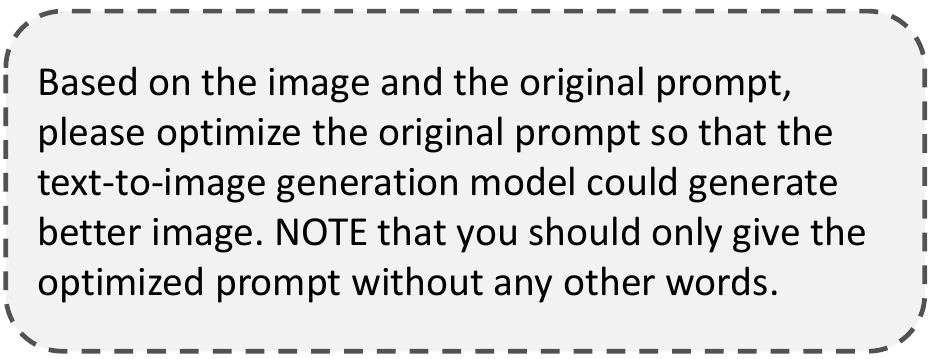}}
    \caption{The system prompt for prompt engineering (PE) with the initial prompt and image as the input.}
    \label{fig:promptlist}
\end{figure}

\begin{figure}[ht]
    \centering
      \scalebox{0.9}{
    \includegraphics[width=\linewidth]{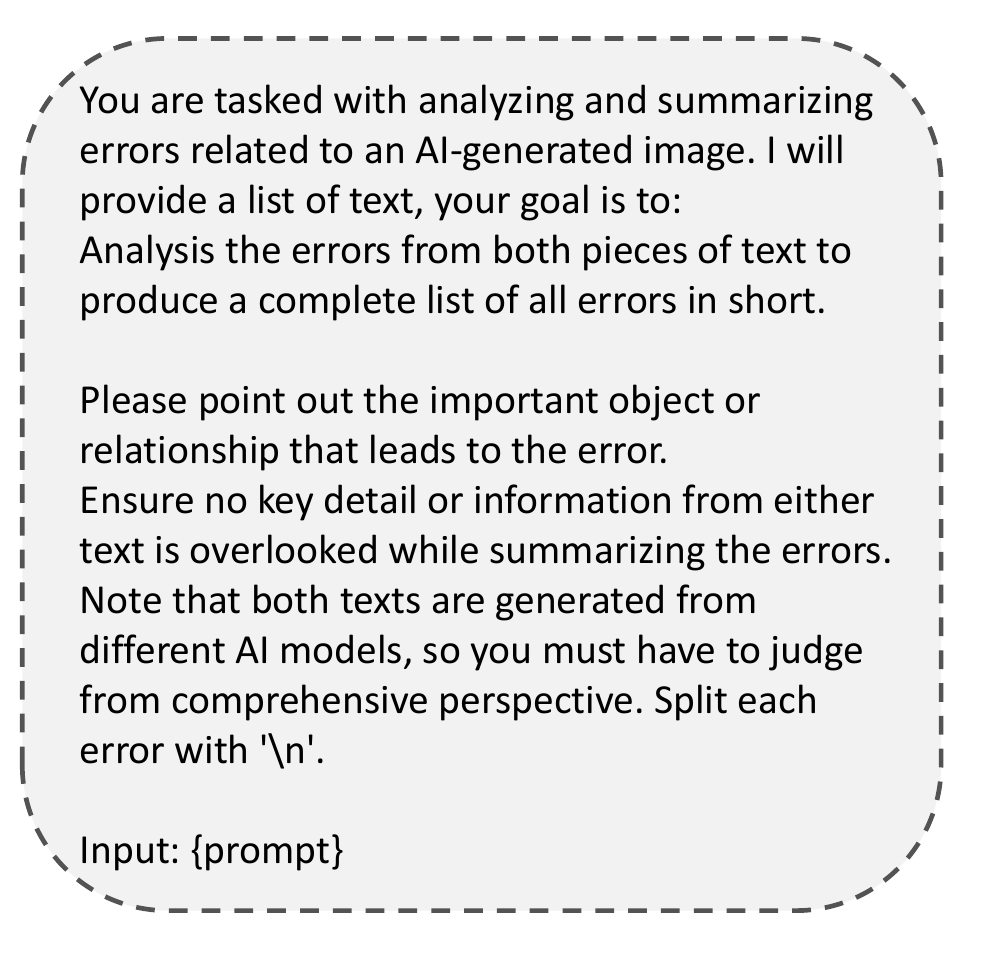}}
    \caption{The system prompt to summarize and explain the reasons for the MLLM agent rating score.}
    \label{fig:feed_back_reasons}
\end{figure}

\begin{figure}[ht]
    \centering
      \scalebox{0.9}{
    \includegraphics[width=\linewidth]{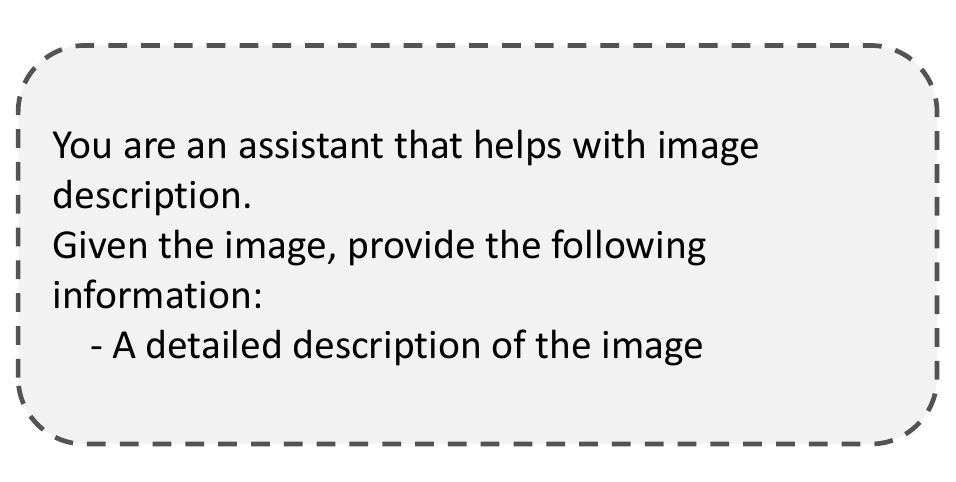}}
    \caption{The system prompt for generating the corresponding caption of the image.}
    \label{fig:image_detail_description}
\end{figure}

\begin{figure*}[htbp]
    \centering
      \scalebox{0.9}{
    \includegraphics[width=\linewidth]{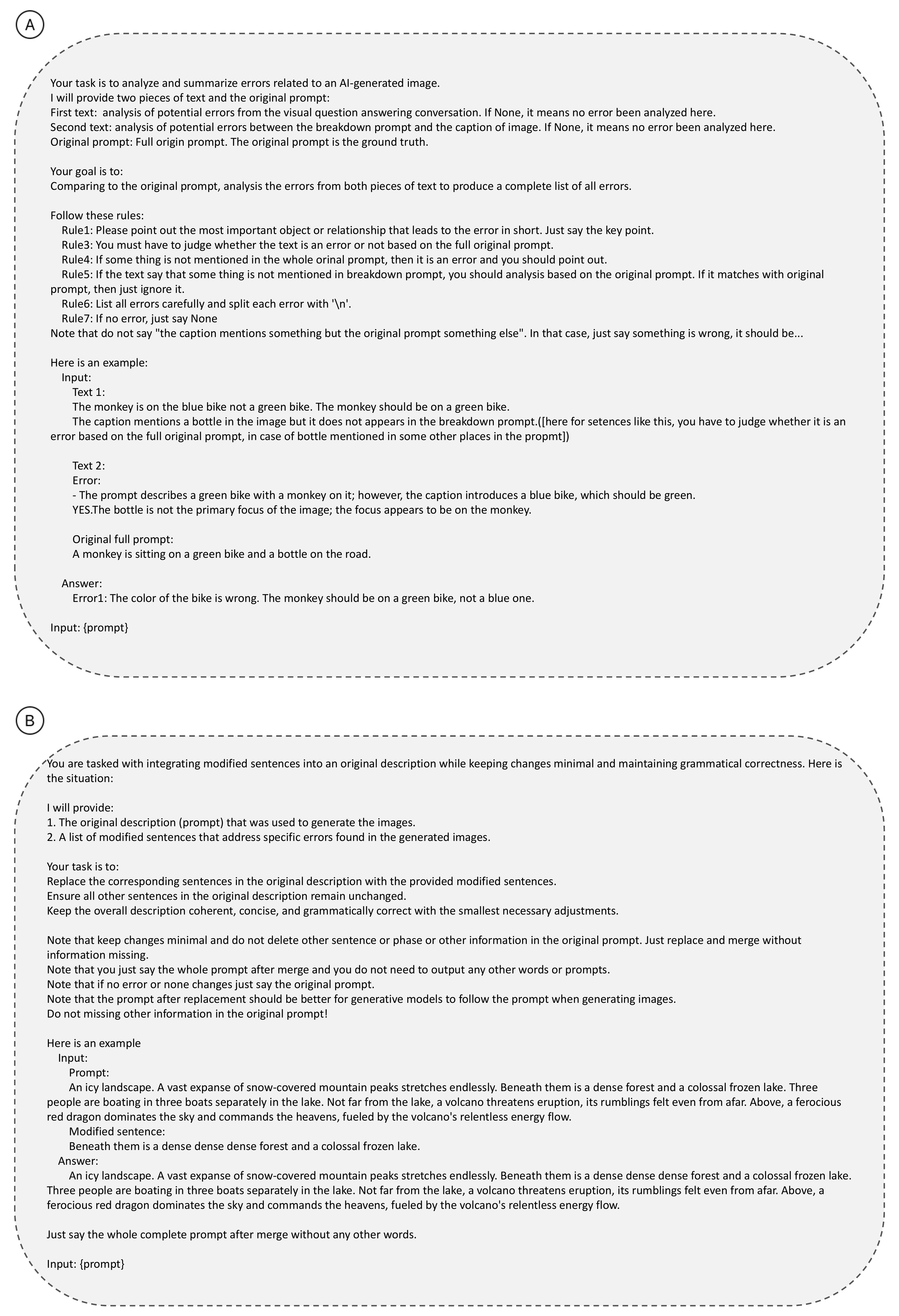}}
    \caption{The system prompt for integration error analysis (A), which combines and verifies the error analysis from VQA-based methods and caption-based analysis, and the instructions for the branch-merge agent for merging the modifications into the original prompt (B).}
    \label{fig:combine+merge}
\end{figure*}

\begin{figure*}[htbp]
    \centering
      \scalebox{0.9}{
    \includegraphics[width=\linewidth]{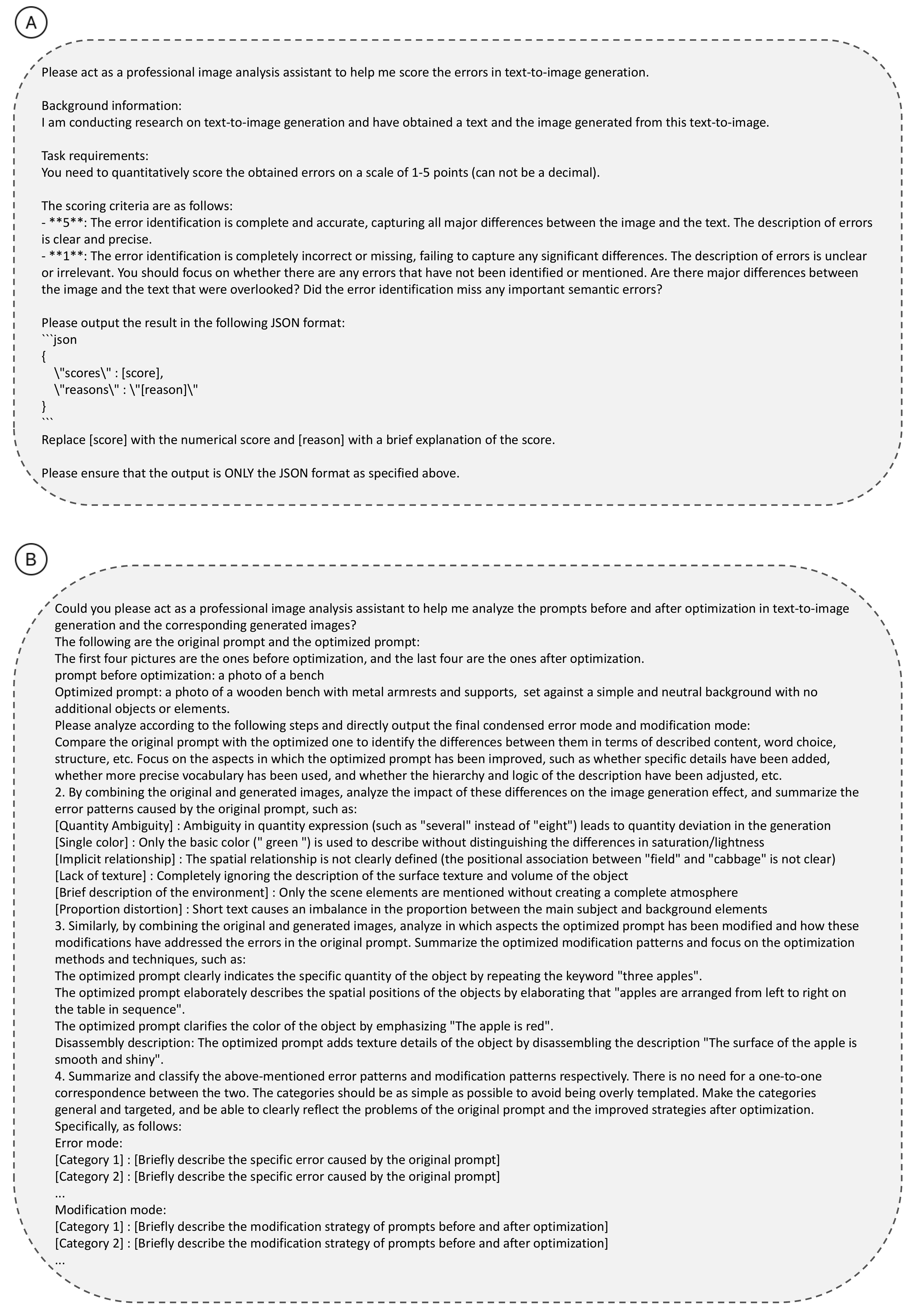}}
    \caption{The system prompt designed for evaluating the accuracy and coverage of error analysis (A), and the instructions to summarize the systematic patterns of errors and optimization strategies (B). }
    \label{fig:ErrorScore+Pattern}
\end{figure*}

\begin{figure*}[htbp]
    \centering
      \scalebox{0.9}{
    \includegraphics[width=\linewidth]{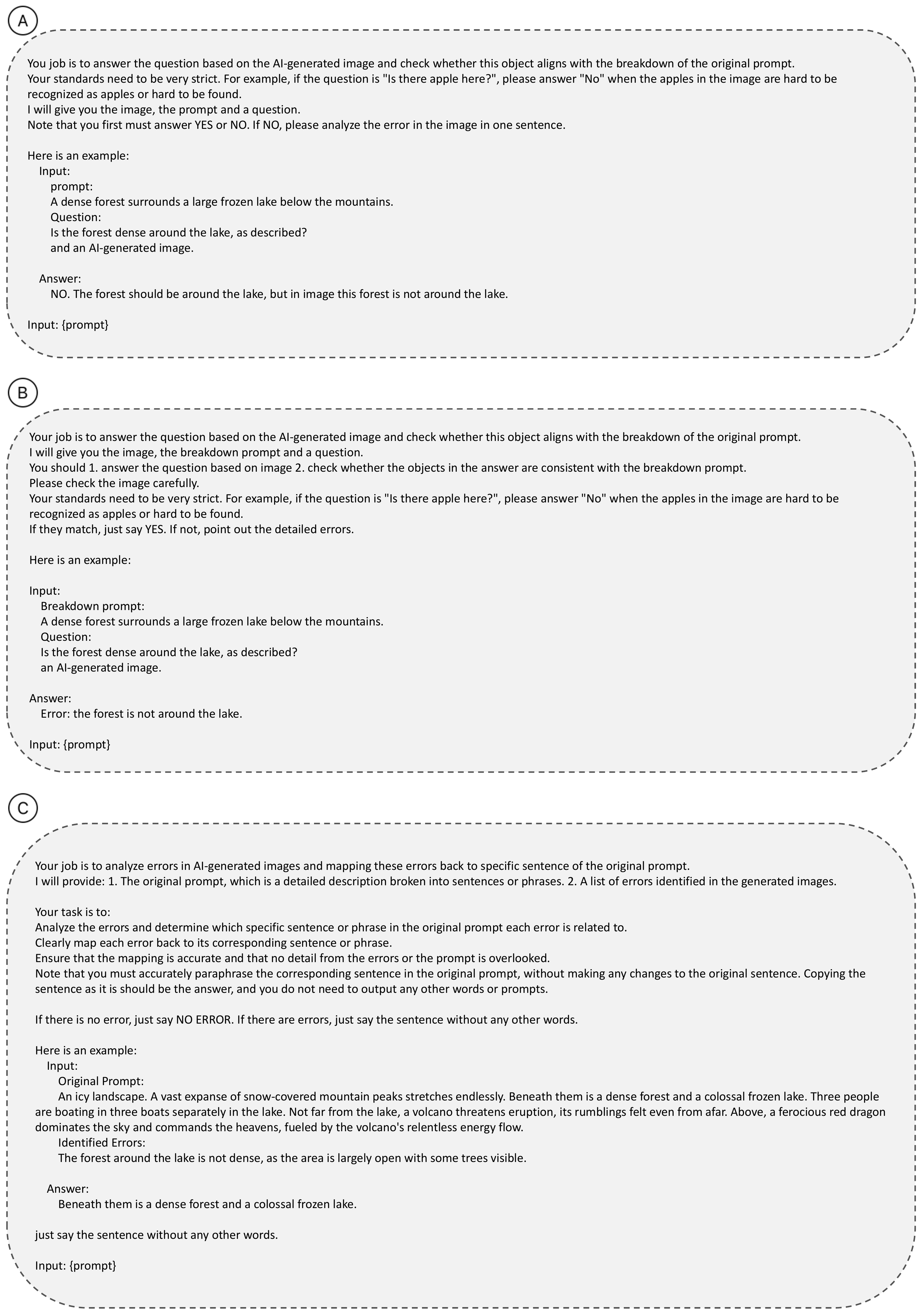}}
    \caption{The system prompt for the VQA module in MLLM scorer (A), the VQA-based error detection (B), and the error mapping (C).}
    \label{fig:qa4+qa+FindBug}
\end{figure*}

\begin{figure*}[htbp]
    \centering
      \scalebox{0.9}{
    \includegraphics[width=\linewidth]{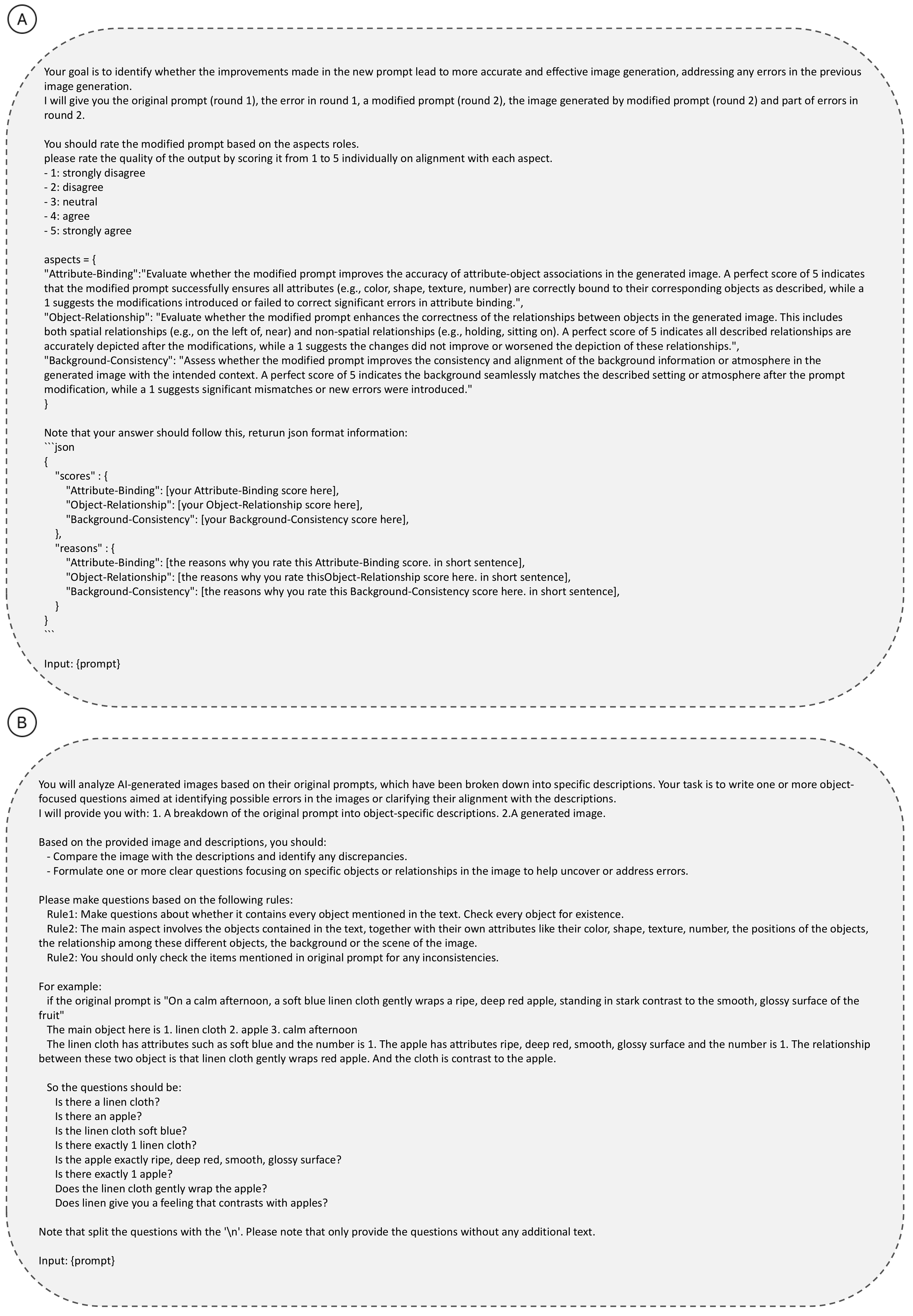}}
    \caption{The rules and strategy for rating the generated images (A) and the structural output in JSON format.
    B represents how we generate the question centered on the object.}
    \label{fig:RateFeedback+Ques}
\end{figure*}

\begin{figure*}[htbp]
    \centering
      \scalebox{0.9}{
    \includegraphics[width=\linewidth]{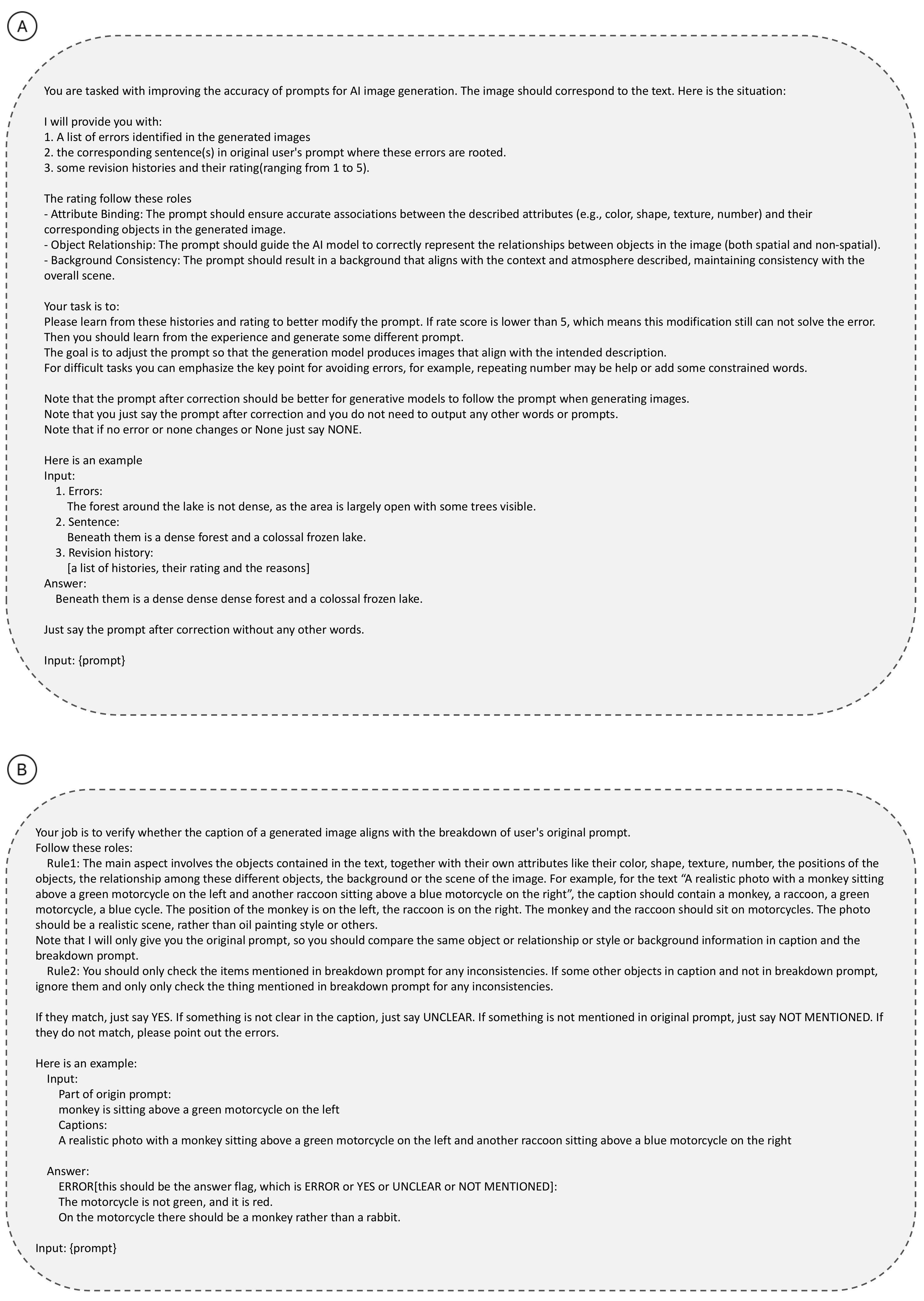}}
    \caption{The system prompt for prompt refinement based on the prompt, image, error analysis, error mapping and revision history, including rate score, feedback, and modified history (A).
    And B refers to the system prompt used for caption-based error detection.}
    \label{fig:RefineWithHistory+CheckCaps}
\end{figure*}

In this section, we provide the system prompt used in GenPilot to guide the agent.
Figure \ref{fig:promptlist} represents the system prompt we use for prompt engineering (PE) in the experiment.
PE takes the original image and prompt as the input and generates the optimized prompts.

In Figure \ref{fig:feed_back_reasons}, we design the instruction for the memory module to store the summary of the detailed errors that occurred, offering a comprehensive reference for the next iteration.

Figure \ref{fig:image_detail_description} is the system prompt we use to generate the detailed descriptive caption of the image.

In Figure \ref{fig:combine+merge}, part a represents the instructions for the error integration agent to verify and summarize the errors. 
The agent will produce a complete list of errors, including patterns and details.
Part B in this figure plays the role of the branch merge agent to combine the modified sentence into the complete prompt.

The instructions for GPT-4o to summarize the error pattern and the refinement pattern are shown in Figure \ref{fig:ErrorScore+Pattern}, part B.
The system prompt in A in the Figure \ref{fig:ErrorScore+Pattern} is used to rate the accuracy and coverage of VQA-based, caption-based, and integration results.

Following the sequence of A, B, and C, Figure \ref{fig:qa4+qa+FindBug} shows the prompt used for the VQA in MLLM scorer, the VQA in error detection, and the error mapping.

In Figure \ref{fig:RateFeedback+Ques}, we provide the instructions for MLLM rating (A) and question list generation (B).

Figure \ref{fig:RefineWithHistory+CheckCaps} shows the system prompt for prompt refinement agent (A) and caption-based error detection (B).

\section{Detailed Example on Comparison for Error Analysis Methods}
\label{errorcase}
\begin{figure}[htbp]
    \centering
    \includegraphics[width=\linewidth]{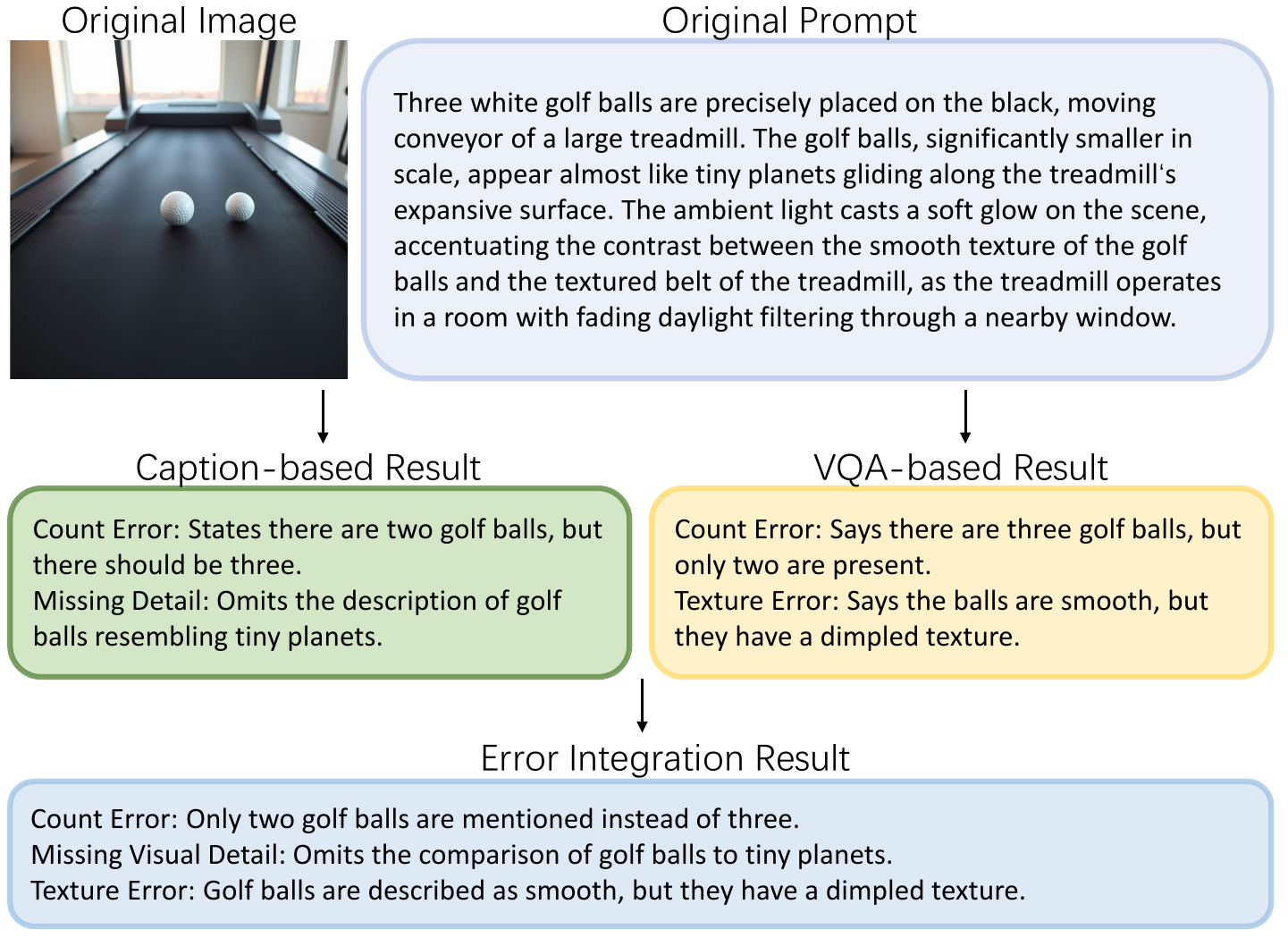}
    \caption{An example that compares the error analysis from the VQA-based method, the caption-based method, and GenPilot.
    According to the original prompt, the inconsistencies are the number, the texture, and the details of golf balls.
    VQA-based method misses the details errors and the caption-based method ignores the texture errors.
    Based on these two analyses, GenPilot is able to perform accurate error analysis.}
    \label{fig:error_showcase}
\end{figure}

As shown in Figure \ref{fig:error_showcase}, GenPilot takes advantage of both methods and verifies each result to generate a complete error analysis.

\section{More Results on DPG-bench}
\label{MoreResultsonDPG-bench}
\begin{figure*}[htbp]
    \centering
      \scalebox{0.8}{
    \includegraphics[width=\linewidth]{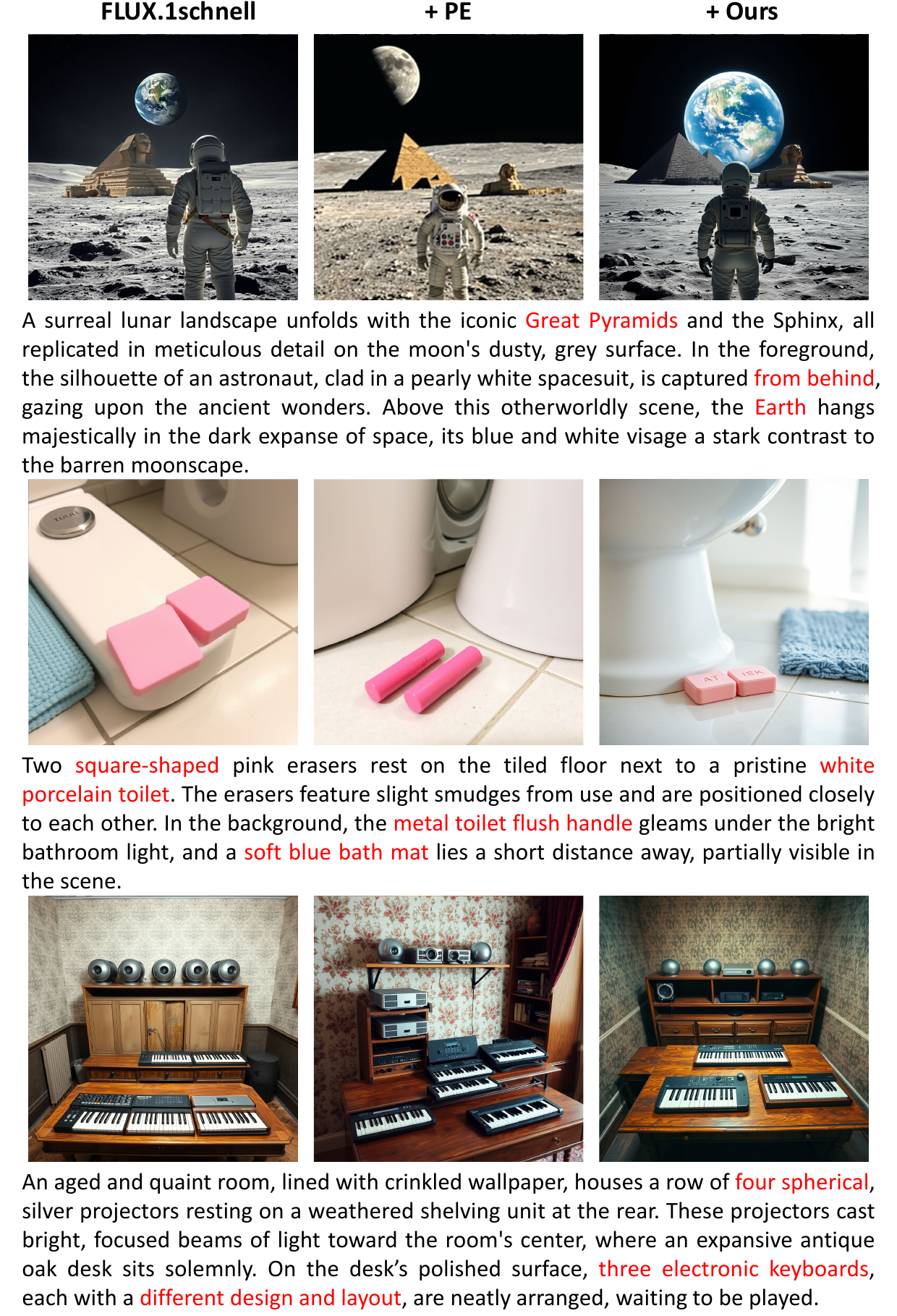}
    }
    \caption{Qualitative results for complex and long prompts on DPG-bench challenging dataset compared to PE and FLUX.1 schenell.
    GenPilot exhibits superior faithfulness to the detailed textual description, for example ``from behind'' and ``metal toilet flush handle'' can be accurately generated with the test-time prompt optimization. }
    \label{fig:flux_dpg_more}
\end{figure*}

In this section, more results conducted on the DPG-bench are illustrated.
As shown in Figure \ref{fig:flux_dpg_more}, we compare the FLUX.1 schnell with PE-optimized and GenPilot-optimized images.
For the first row, the main objects in the original prompt are the Pyramids, the Sphinx, an astronaut, and Earth.
Our method clearly renders the iconic Great Pyramids, the Sphinx, the astronaut from behind, and a vividly contrasting Earth, while PE provides an astronaut from the front, and the FLUX image mistakenly combines the Pyramids and the Sphinx together.
In the second row, our approach successfully generates ``two square-shaped pink erasers'' next to a toilet, compared to the square erasers on the toilet in the FLUX image and the tube-shaped erasers in the PE image.
Moreover, PE image misses the blue bath mat and the handle in the background.
Finally, in the challenging prompt of an aged room with multiple projectors and keyboards in the third row, GenPilot accurately generates 4 spherical, silver projectors, in contrast to the 5 and 2 in FLUX image and PE image.
These qualitative comparisons in Figure \ref{fig:flux_dpg_more} demonstrate the superior ability of GenPilot to interpret complex prompts for enhanced image generation.
Our approach accurately renders distinct objects with their specified attributes, correct spatial relationships, and the precise number, revealing the effectiveness and potential of GenPilot to improve image quality in text-to-image synthesis.
\begin{figure*}[htbp]
    \centering
  \scalebox{0.8}{
    \includegraphics[width=\linewidth]{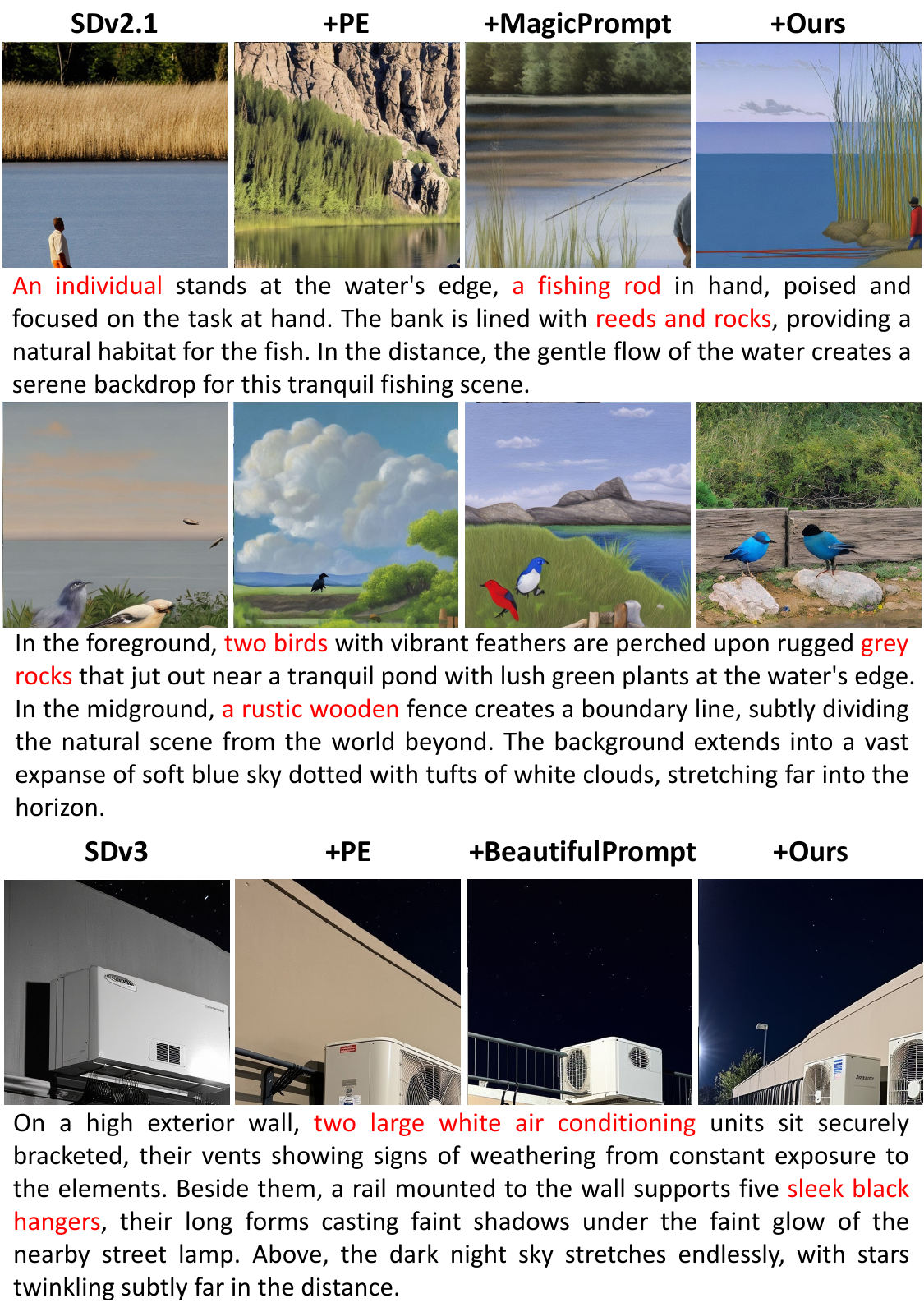}
    }
    \caption{The additional examples on DPG-bench challenging dataset with SDv2.1 in the first and second row, and with SD3 on the last row.
    For comparison, we choose the higher baseline method from BeautifulPrompt and MagicPrompt.
    The results highlight the superiority of GenPilot in accurately rendering complicated scenes compared to generative models and other enhancement methods.
    }
    \label{fig:append_sd}
\end{figure*}

\begin{figure*}[htbp]
    \centering
      \scalebox{0.8}{
    \includegraphics[width=\linewidth]{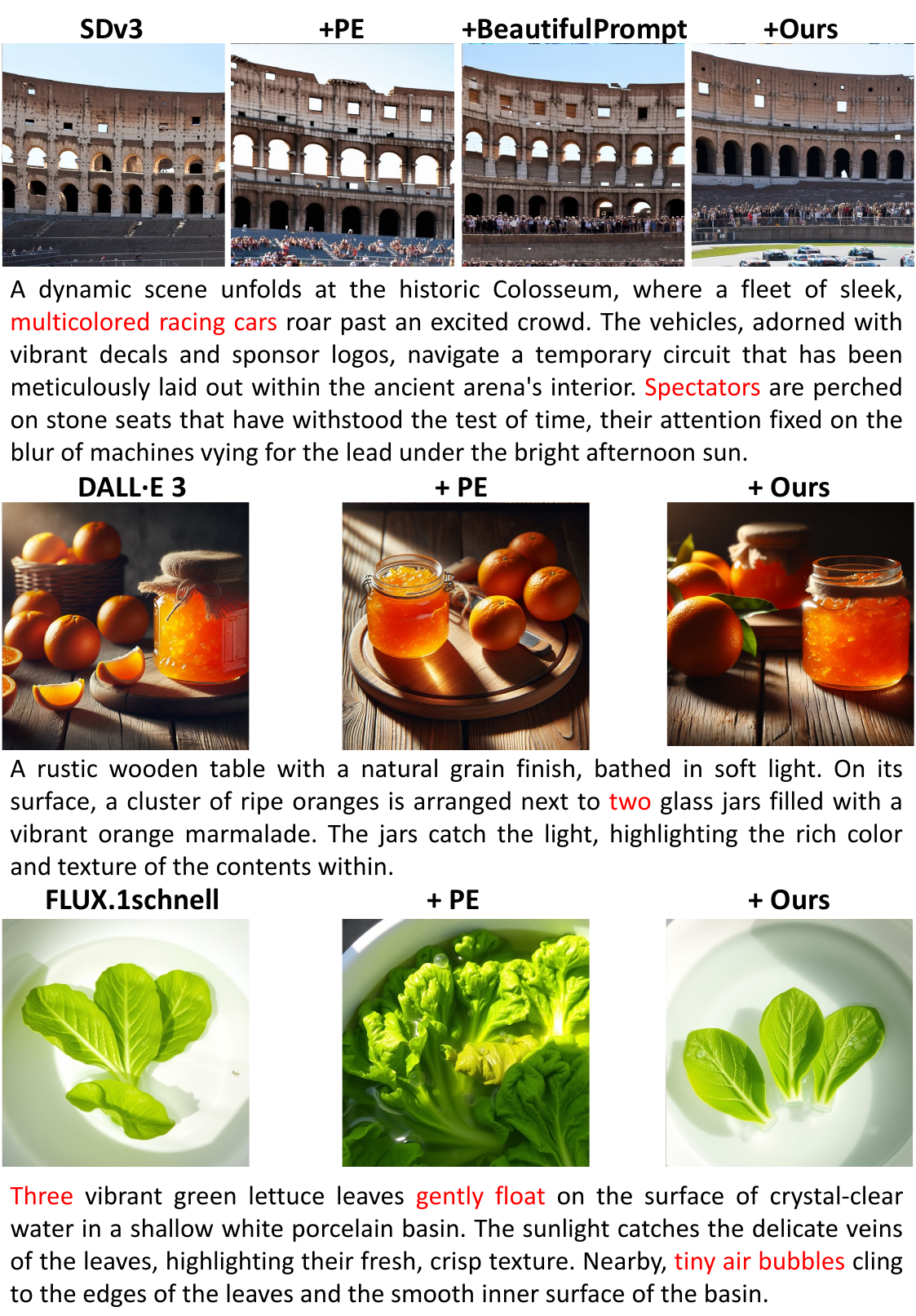}
    }
    \caption{More qualitative results on DPG-bench challenging dataset with SD3 in the first row, DALL-E 3 on the second row, and FLUX.1 schnell on the last row.
    The results clearly demonstrate the significant advantages of our method over the FLUX.1 schnell and the PE method.
    Specifically, our approach accurately renders key details from the prompt, such as ``three''.
    }
    \label{fig:appendix_sd3_dpg}
\end{figure*}

\section{More Results on GenEval}
\label{MoreResultsonGenEval}
\begin{figure*}[htbp]
    \centering
      \scalebox{0.8}{
    \includegraphics[width=\linewidth]{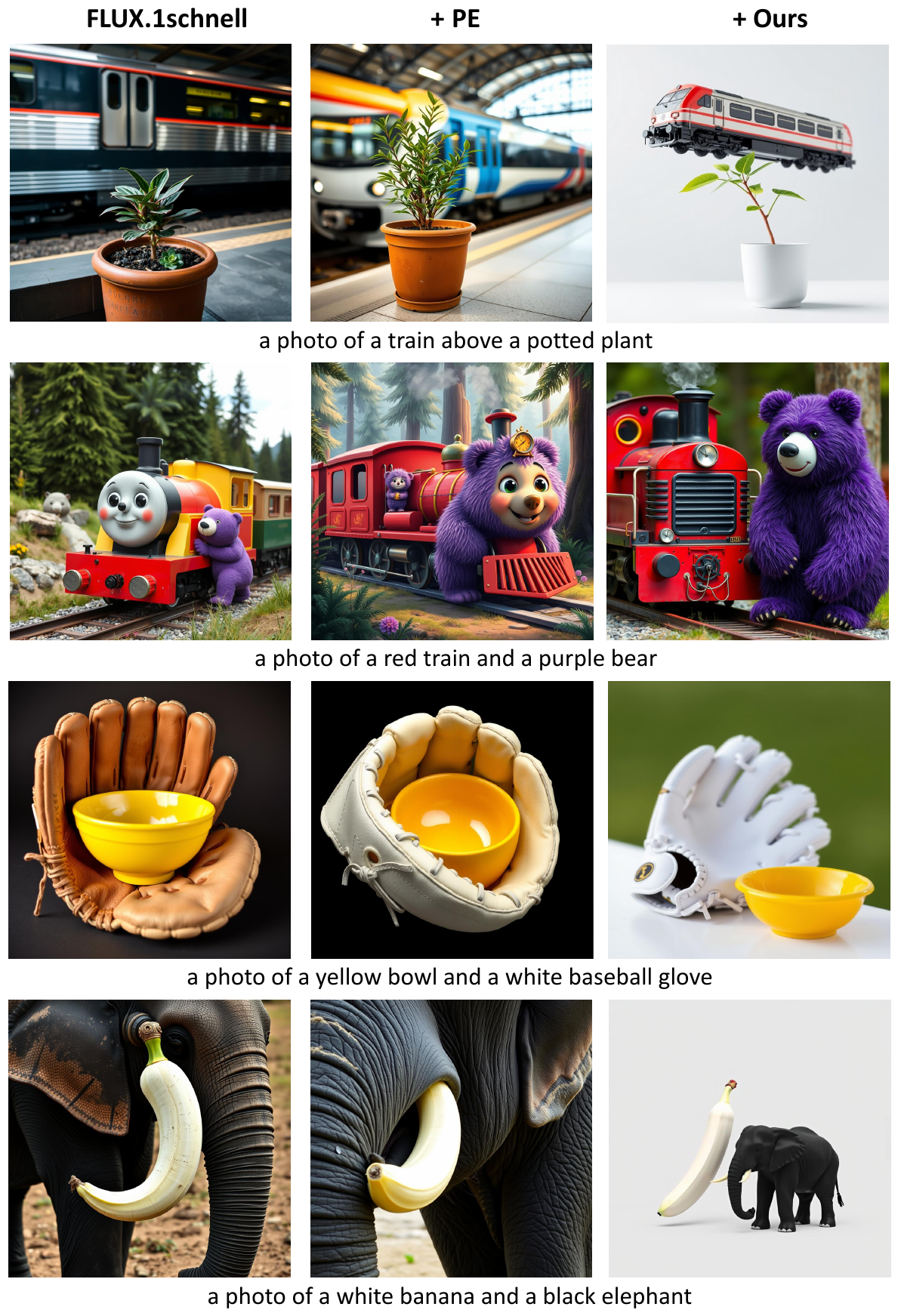}
    }
    \caption{The qualitative results on GenEval with FLUX.1 schnell, FLUX.1 schnell with PE, and FLUX.1 schnell with GenPilot.
    Our system accurately synthesizes the unrealistic image, demonstrating the significant superiority of our method in understanding and generating images.}
    \label{fig:flux_gen_2}
\end{figure*}

\begin{figure*}[htbp]
    \centering
      \scalebox{0.8}{
    \includegraphics[width=\linewidth]{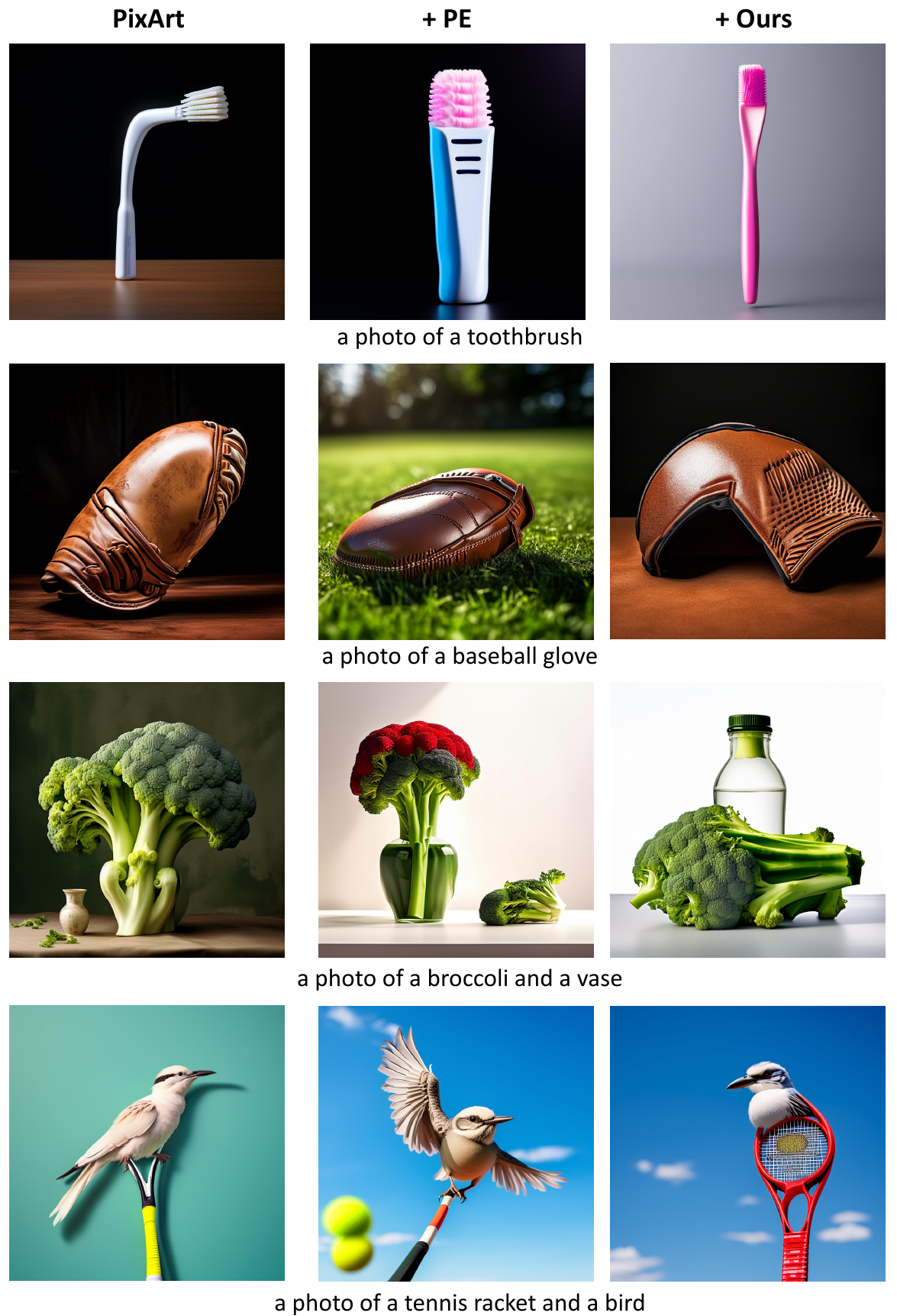}
    }
    \caption{Additional examples on GenEval with PixArt-$\alpha$ and enhancement methods PE and GenPilot.
    The results reveal the advantages of dealing with the details, such as the baseball glove of GenPilot.}
    \label{fig:pixart_gen_1}
\end{figure*}

In this section, we provide more qualitative experimental results on the GenEval benchmark, as shown in Figure \ref{fig:flux_gen_2} and Figure \ref{fig:pixart_gen_1}.
Though Flux.1 schnell and PixArt-$\alpha$ have achieved relatively great performance, sometimes they may fail, such as in the unrealistic ones and position-related prompts.
In Figure \ref{fig:flux_gen_2}, when the prompt describes an uncommon scene, ``a photo of a train above a potted plant'', Flux.1 schnell generates an image of a train behind a plant, which is consistent with real-world principles.
With GenPilot, Flux.1 schnell can accurately generate an unreal scene with a train floating above a plant.

PixArt-$\alpha$ in Figure \ref{fig:pixart_gen_1} is not skilled in drawing shapes and details, especially for the combination of multiple objects.
In contrast, with GenPilot, PixArt-$\alpha$ is capable of generating specific details, for example, the image in the second row of a baseball glove.

The qualitative results highlight the effectiveness and capability of seamlessly applying to various models.

% \section{Detailed Example on Comparison for Error Analysis Methods}
% \label{errorcase}
% \begin{figure}[htbp]
%     \centering
%     \includegraphics[width=\linewidth]{img/error_showcase.pdf}
%     \caption{An example that compares the error analysis from the VQA-based method, the caption-based method, and GenPilot.
%     %
%     According to the original prompt, the inconsistencies are the number, the texture, and the details of golf balls.
%     %
%     VQA-based method misses the details errors and the caption-based method ignores the texture errors.
%     %
%     Based on these two analyses, GenPilot is able to perform accurate error analysis.}
%     \label{fig:error_showcase}
% \end{figure}

% As shown in Figure \ref{fig:error_showcase}, GenPilot takes advantage of both methods and verifies each result to generate a complete error analysis.

\section{Detailed Pattern on Error and Optimization Analysis}
\label{DetailedPatternonErrorandOptimizationAnalysis}
In this section, we list the patterns of errors and the refinement strategy summarized by GPT-4o based on the original prompt and optimized prompt.
The system prompt for GPT-4o can be found at Appendix \ref{SystemPromptTemplate}.
We release 35 patterns and their corresponding refinement strategy, along with cases for better understanding.

\underline{Quantity Errors}:
Quantity Errors refer to the number of objects in the generated image that does not match the description in the prompt. 
To address this issue, the optimized prompt employs a strategy of repeating quantity keywords and incorporates the adverbs ``exactly'' and ``precisely'' to enhance precision.
For example, the original prompt did not guarantee the correct depiction of exactly eight chairs. 
The optimized prompt emphasizes the exact number of ``eight chairs'' and uses ``exactly'' to reinforce the precision of the quantity, thereby ensuring that the generated image accurately reflects the specified number of objects.

\underline{Spatial Positioning Errors}:
Spatial Positioning Errors arise when objects in the generated image are placed incorrectly relative to one another.
The optimized prompt addresses this by introducing a more systematic approach to spatial description. It explicitly defines objects' coordinates, angles, and distances to other objects within a three-dimensional framework.
For example, the original prompt caused errors in the depiction of the boy's position relative to the woman, resulting in inconsistencies with the intended positioning.
The optimized prompt clarifies spatial positions with terms like ``precisely'' and ``directly behind'' to reduce ambiguity and ensures that spatial relationships are conveyed unambiguously, thereby minimizing spatial positioning errors and eliminating inconsistencies in the generated image.

\underline{Texture Errors}:  
Texture Errors happen when the surface textures of objects in the generated image do not match real-world expectations or appear missing.  
The optimized prompt tackles this issue by introducing more detailed texture descriptions and emphasizing them.  
For example, the original prompt failed to highlight the frosty texture on the boards, which is inadequately visible.  
The optimized prompt provides more detailed descriptions of the texture and repeatedly emphasizes the frosty texture on both ice and boards to correct texture visibility errors and make the generated image more realistic.

\underline{Color Errors}:  
Color Errors mean the colors of objects in the generated image deviate from the specified requirements.  
The optimized prompt introduces a more systematic approach to color description by incorporating precise color terminology and describing colors across multiple dimensions such as hue, brightness, and saturation.  
For example, the original prompt's lack of specificity in defining the pear's color resulted in variations and potential color mismatches in the output.  
To address this, the optimized prompt employs exact color references like ``Pantone 376C'' to specify the pear's color, thereby reducing ambiguity and enhancing color accuracy in the generated image.

\underline{Shape Errors}:  
Shape Errors occur when the shapes of objects in the generated image do not meet the requirements or are illogical.  
The optimized prompt tackles this issue by repeatedly emphasizing the unique shape of the object and adding detailed descriptions.  
For example, the glasses on the horse were not clearly differentiated in terms of color and shape from the original prompt.  
The optimized prompt provides a clearer distinction for the types of glasses by detailing their specific colors and frame shapes through expanded descriptions, thereby enhancing the accuracy and logic of the object's shape in the generated image.

\underline{Proportion Errors}:  
Proportion Errors refer to the scale and size of objects in the generated image are imbalanced or illogical.  
The optimized prompt addresses this by providing detailed descriptions of object proportions and introducing specific measurement references.  
For example, the original prompt failed to effectively depict the size relationship between the oversized blue rubber ball and the net and hoop.  
The optimized prompt emphasizes the impossibility of the ball passing through the hoop by enhancing the description of the ball's oversized nature, thereby ensuring a more realistic representation of proportions in the generated image.

\underline{Action or Pose Errors}:  
Action or Pose Errors occur when the movements or postures of figures or animals in the generated image do not align with the description or logical expectations.  
The optimized prompt addresses this by incorporating detailed action descriptions and emphasizing dynamic balance.  
For example, the original prompt resulted in ambiguities related to spatial relationships between body parts, especially in arm and leg positioning, which affected the sense of balance.  
The optimized prompt utilizes specific descriptions to define spatial relationships and emphasizes precise alignment and balance, thereby enhancing the overall dynamic and harmonic posture in the generated image.

\underline{Scene Element Omissions}:  
Scene Element Omissions occur when key components of a scene are missing or underrepresented in the generated image.  
The optimized prompt solves this by explicitly listing all critical elements required in the scene and reiterating their quantity and spatial relationships.  
For example, the original prompt mentioned tools and metal racks but failed to highlight their prominence, resulting in a minimalist scene that deviated from the intended complexity.  
The optimized prompt explicitly lists elements like ``tools'' and ``metal racks'' through repetition, ensuring they are visually emphasized and properly positioned, thereby enriching the scene and aligning it with the detailed description provided

\underline{Extraneous Scene Elements}:  
Extraneous Scene Elements arise when the generated image includes objects or components not specified in the prompt.  
The optimized prompt addresses this issue by explicitly excluding unnecessary elements and emphasizing their absence.  
For example, the original prompt failed to specify the absence of other furniture or objects in the room, leading to the inclusion of unintended elements.  
The optimized prompt distinctly stated the absence of other elements like benches or lighting in the room, thereby preventing superfluous additions and ensuring the scene remains faithful to the intended description.

\underline{Indistinct Background Errors}:  
Indistinct Background Errors mean the background details in the generated image are unclear or underdeveloped.  
The optimized prompt solves this by explicitly enumerating background elements and emphasizing their characteristics, positions, and spatial relationships relative to the foreground.  
For example, the original prompt's vague description of the evening sky and surrounding foliage resulted in inconsistent or underdeveloped background details.  
The optimized prompt added precise descriptions of elements like ``large, reflective aviator sunglasses'' and the cat's ``small, furry face'' ensuring these features are clearly generated while also detailing the background to enhance overall image coherence.

\underline{Lighting Errors}:
Lighting Errors arise when the generated image features lighting that does not align with the intended direction or intensity as described in the prompt.
The optimized prompt addresses this by explicitly defining the light source, direction, intensity, color, and its interplay with objects in the scene.
For example, the original prompt failed to effectively capture the interaction between light and mist, which is critical for creating a misty atmosphere.
The optimized prompt greatly enhances the accuracy of the lighting elements by specifically defining the light's origin, direction, intensity, color, and interaction with mist.

\underline{Shadow Errors}:  
Shadow Errors happen when the position and shape of shadows in the generated image don't match the light source and objects.  
The optimized prompt tackles this issue by clearly specifying the light source, direction, intensity, and the material and shape of objects.  
For instance, the original prompt's lighting didn't consistently emphasize the bristles or cast long shadows, leading to inaccurate shadow patterns.  
The optimized prompt highlights the monitor's soft glow as the main light source for highlighting the bristles and casting shadows, enhancing shadow depiction accuracy, and ensuring light sources, objects, and their shadows are consistent in the generated image.

\underline{Reflection Errors}:  
Reflection Errors occur when the reflection on object surfaces does not comply with physical laws.  
The optimized prompt addresses this by strengthening the description of the reflection process and detailing the light source's origin, direction, intensity, and the object's material and surface properties.  
For instance, the original prompt failed to effectively capture the reflection of lighting on the desk's surface due to a lack of emphasis on reflective qualities.  
The optimized prompt enhances the description of lighting reflection by using phrases like ``clearly reflecting the soft glow'' thereby improving clarity on reflective surfaces and ensuring the generated image adheres to physical reflection principles.

\underline{Object Blurriness}:  
Object Blurriness happens when object outlines and details are unclear in the generated image.  
The optimized prompt addresses this by emphasizing clear contours and layered details, introducing terms like ``sharpness'' and ``high resolution'' while providing multi-level descriptions of local details.  
For example, the original prompt resulted in a blurred depiction of the anime character’s facial features.  
The optimized prompt emphasizes ``ultra-high-definition rendering'' and specifies details like ``distinct eyelash strands'' and ``subtle skin pores visible under studio lighting'' to ensure clarity in both macro and micro details of the object.

\underline{Style Errors}:  
Style Errors arise when the overall style or specific elements in the generated image deviate from the intended aesthetic.  
The optimized prompt addresses this by introducing stylized keywords and specifying style characteristics such as line thickness, color saturation, and lighting treatment, all while emphasizing stylistic uniformity.  
For example, the original prompt led to inconsistencies in the steampunk style of the clockwork mechanism.  
The optimized prompt specifies features like “hyper-detailed brass gears with visible rivets” and “soft Edison bulb illumination” to enforce stylistic coherence across all components, ensuring a unified visual style in the generated image.

\underline{Material Errors}:  
Material Errors happen when the generated image inaccurately represents the material properties of objects.  
The optimized prompt addresses this by explicitly specifying the physical attributes of materials, such as roughness, glossiness, and transparency.  
For example, the original prompt failed to render the metallic texture of the samurai armor.  
The optimized prompt uses precise material descriptors like “matte blackened steel with brushed titanium accents” to refine material fidelity, ensuring the generated image accurately reflects the intended texture and finish.

\underline{Composition Errors}:  
Composition Errors arise when the layout of the scene in the generated image does not meet the requirements or defies common sense.  
The optimized prompt resolves this issue by combining composition keywords with quantified object placement and proportion, and by clarifying the hierarchical relationship between the main subject and the background.  
For example, the original prompt resulted in an unbalanced composition with the main subject positioned at the edge of the frame.  
The optimized prompt specifies ``precise frame composition with the subject centered at the golden ratio point'' to achieve harmonic visual balance, ensuring the layout aligns with the intended design principles.

\underline{Interaction Errors}:  
Interaction Errors occur when the relationships between objects in the generated image are incorrectly portrayed.  
The optimized prompt addresses this by using emphasis and contrast to enhance the description of interaction details, ensuring vivid and accurate depictions of how objects interact.  
For example, the faint trail of damp grass left on the ball as it moves was entirely missing in the original prompt.  
The optimized prompt includes a clearer depiction of the interaction between the damp grass and the rolling baseball, ensuring the faint trail is distinctly noticeable and the interaction between the two elements is portrayed realistically.

\underline{Ambiguous Object States}:  
Ambiguous Object States occur when the condition or status of objects in the generated image is unclear.  
The optimized prompt addresses this by explicitly defining the specific state of objects, such as motion, power status, or deformation, and incorporating dynamic descriptions.  
For example, the original prompt led to ambiguity in whether the lamp was on or off.  
The optimized prompt specifies ``the lamp is in an on state with warm light'' to clarify its operational status, ensuring the generated image accurately reflects the intended state of the object.

\underline{Object Fusion Errors}:
Object Fusion Errors happen when multiple objects in the generated image are incorrectly merged together.
The optimized prompt addresses this by emphasizing the independence and boundaries of objects, employing clear separation descriptions.
For example, the original prompt caused the cat and the dog to merge into a single indistinct shape.
The optimized prompt specifies ``next to'' and enforces ``visible fur texture differentiation'' to maintain their individual identities, preserving the distinctness of each object in the generated image.

\underline{Emphasis Errors}:  
Emphasis Errors occur when elements that should be highlighted in the generated image are not sufficiently emphasized.  
The optimized prompt addresses this issue by incorporating emphasis keywords such as ``highlight'' and ``emphasize'', combined with contrastive descriptions to draw attention to focal points.  
For example, the original prompt failed to highlight the majestic appearance of the dragon.  
The optimized prompt emphasizes ``the dragon’s scales gleaming with iridescent hues'' to ensure it stands out as the focal point, thereby enhancing the visual impact and ensuring the intended elements are prominently featured in the generated image.

\underline{Atmospheric Mismatch Errors}:  
Atmospheric Mismatch Errors occur when the generated image fails to align with the intended mood or atmosphere described in the prompt.  
The optimized prompt addresses this by incorporating explicit atmospheric keywords like ``mood'' and ``atmosphere'' alongside detailed descriptions of environmental elements such as lighting, color tones, and specific details.  
For example, the original prompt failed to create the intended mysterious forest atmosphere.  
The optimized prompt emphasizes ``a dark and mysterious atmosphere with fog swirling around ancient tree roots'' and specifies ``dappled moonlight filtering through dense branches with a cool blue tone'' to enhance the intended mood, ensuring the generated image effectively conveys the desired atmosphere.

\underline{Cluttered Background Errors}:  
Cluttered Background Errors arise when background elements in the generated image are excessive or disorderly, detracting from the main focus.  
The optimized prompt addresses this by defining a neutral and clean background and imposing restrictions on background elements.  
For example, inadequate details in the original prompt led to distracting background elements that interfered with the scene's focus.  
The optimized prompt defined a neutral and uncluttered background to emphasize the piano and bench, preventing distractions and ensuring the main subjects remain the focal point in the generated image.

\underline{Partial Object Generation}:  
Partial Object Generation happens when parts of objects in the generated image are missing.  
The optimized prompt addresses this by detailing the object's overall structure and each part's specifics, clarifying the connections between parts, and repeatedly emphasizing the object's completeness.  
For example, the original prompt caused the generation of a bicycle with a missing rear wheel.  
The optimized prompt specifies the ``complete structure of a bicycle with two wheels'' and repeatedly emphasizes that ``all components, including handlebars, seat, pedals, and both wheels, are fully intact and firmly attached'' to ensure no part is omitted in the generated image.

\underline{Object Occlusion Errors}:  
Object Occlusion Errors occur when key parts of objects in the generated image are inappropriately blocked by other elements.  
The optimized prompt addresses this by explicitly defining the spatial hierarchy between objects and emphasizing the visibility of critical items.  
For example, the original prompt caused the woman's face to be partially obscured by the vase in the foreground.  
The optimized prompt specifies the ``woman positioned in the foreground with a clear, unobstructed view of her face'' and adjusts the spatial arrangement by stating ``the vase placed behind the woman'' ensuring key elements remain visible and the intended focus is maintained in the generated image.

\underline{Unwanted Brand Elements}:  
Unwanted Brand Elements occur when brand logos or identifiers appear inappropriately in the generated image.  
The optimized prompt addresses this by explicitly stating the exclusion of any brand characteristics and emphasizing their absence.  
In this case, the original image features undesired brand symbols such as ``NEFE'' on the paintbrush, which were not specified in the original prompt.  
The optimized prompt explicitly excludes brand names or symbols, leading to cleaner results without unwanted visual elements.

\underline{Temporal Ambiguity Errors}:  
Temporal Ambiguity Errors occur when the time setting in the generated image is unclear or inaccurately represented.  
The optimized prompt addresses this by explicitly specifying the exact time point or time period to eliminate ambiguity.  
For example, the original prompt inadequately linked the scene to a clear midnight context, creating ambiguity regarding the setting.  
In the optimized prompt, midnight context details were reinforced with references to the moon’s alignment, object illumination, and atmospheric serenity, ensuring a precise and unambiguous temporal setting in the generated image.

\underline{Seasonal Element Errors}:  
Seasonal Element Errors occur when elements related to seasons in the generated image are illogical or inconsistent.  
The optimized prompt addresses this by explicitly specifying the exact season and detailing natural characteristics, climate conditions, and typical activities associated with that season.  
For example, the original prompt led to confusion over specific items associated with each season, resulting in misplaced elements like pumpkins in spring and summer images.  
The optimized prompt explicitly rejects inappropriate additional objects and emphasizes relevant seasonal motifs and colors, ensuring the generated image accurately reflects the intended season.

\underline{Facial Expression Errors}:  
Facial Expression Errors occur when the facial expressions of characters in the generated image do not align with the intended emotions described in the prompt.  
The optimized prompt addresses this by providing detailed descriptions of facial features and utilizing environmental contrasts to highlight the desired expression.  
For example, the original prompt failed to fully convey the fierce expression of fiery vengeance, particularly in the eyes and mouth area.  
The optimized prompt intensely highlighted key facial features with flames to enhance the skull’s menacing and vengeful expression, ensuring the generated image accurately reflects the intended emotion through detailed facial rendering and environmental emphasis.

\underline{Transparency Errors}:  
Transparency Errors occur when the transparency of objects in the generated image is depicted unreasonably.
The optimized prompt addresses this by emphasizing the transparent effects of objects and providing details on how light refracts and reflects through them, as well as how other objects are reflected.
For example, the astronaut's helmet does not correctly reflect or have transparency showing the lunar landscape. The improved prompt focuses on the helmet's transparency property, ensuring its integration with the lunar landscape.

\underline{Background Inconsistency Errors}:  
Background Inconsistency Errors happen when the style and elements of the background in the generated image are not unified.  
The optimized prompt addresses this issue by emphasizing the need for background consistency and setting an overall style theme.  
For instance, the original prompt failed to ensure the entire scene, including the background, maintained visual coherence.  
The optimized prompt establishes ``a historical 15th-century European setting'' and stresses the importance of keeping the background consistent with this theme, thereby achieving visual harmony in the generated image.

\underline{Contrast Errors}:  
Contrast Errors arise when the overall contrast in the generated image is inappropriate, leading to poorly defined distinctions between elements.  
The optimized prompt resolves this by explicitly specifying the desired contrast and reinforcing related descriptions.  
For example, the lack of emphasis on visual contrast between apples and leaves resulted in less defined distinctions in the images.  
The optimized prompt highlights the contrast between the circular leaves and square apples to improve visual discrepancy, ensuring the generated image reflects the intended clarity and distinction through enhanced contrast.

\underline{Color Disharmony Errors}:
Color Disharmony Errors occur when colors in the generated image clash or fail to create a harmonious visual effect.
The optimized prompt addresses this by emphasizing the need for overall color coordination and providing a harmonious color scheme.
For example, the lack of emphasis on color harmony between elements led to disjointed visual tone representations. By emphasizing color harmony and warm tones, the optimized prompt established better visual and thematic coherence.

\underline{Emotional Tone Errors}:  
Emotional Tone Errors occur when the overall image fails to convey the intended emotional tone.  
The optimized prompt addresses this by elaborately describing the emotional atmosphere and integrating elements like color, lighting, and composition, while emphasizing emotional keywords.  
For example, the original prompt failed to capture the intended sadness in the scene of a solitary figure by the window.  
The optimized prompt counters this by specifying ``a melancholic atmosphere with soft blue and gray tones,'' ensuring the generated image aligns with the desired emotional impact through cohesive use of color and lighting.

\underline{Object Boundary Errors}:  
Object Boundary Errors occur when the boundaries of objects in the generated image are unclear or incomplete.  
The optimized prompt addresses this by detailing the object’s contours, edge characteristics, and contrast with the background, thereby emphasizing clear boundaries.  
For instance, the original prompt resulted in a tree whose edges blended ambiguously with the background, making the boundary unclear.  
The optimized prompt specifies that the tree should have ``crisp, well-defined edges with high contrast against the sky'' ensuring the object stands out distinctly in the generated image.

\end{document}